\pdfoutput=1
\documentclass[journal]{IEEEtran}

\usepackage{amssymb}

\usepackage{graphicx}
\usepackage{subfig}
\usepackage{tabularx}

\usepackage{algorithm2e}

\usepackage{algpseudocode}

\usepackage{amsmath} 
\usepackage{mathtools}
\usepackage{bm}
\usepackage{esvect}

\usepackage{amsmath}
\usepackage{amssymb}
\usepackage{dsfont}

\usepackage{graphicx}
\usepackage{subfig}

\usepackage{floatrow}

\usepackage{scalerel}

\usepackage{balance}
\usepackage{float}

\newcommand{\aggregate}[2]{\underset{#2}{\operatornamewithlimits{#1\ }}}

\newcommand*{\horzbar}{\rule[.5ex]{2.5ex}{0.5pt}}

\makeatletter
\renewcommand*\env@matrix[1][\arraystretch]{%
  \edef\arraystretch{#1}%
  \hskip -\arraycolsep
  \let\@ifnextchar\new@ifnextchar
  \array{*\c@MaxMatrixCols c}}
\makeatother

\SetKwInput{KwInput}{Input}                
\SetKwInput{KwOutput}{Output}              

\usepackage{multirow}
\usepackage[dvipsnames]{xcolor}

\renewcommand\arraystretch{1.7}

\usepackage[T1]{fontenc}
\usepackage[utf8]{inputenc}
\usepackage{setspace}
\usepackage{lipsum}
\usepackage{booktabs}
\usepackage{rotating}
\usepackage{siunitx}
\usepackage[skip=0.1\baselineskip]{caption}
\definecolor{mygray}{gray}{0.4}

\usepackage{float}
\floatstyle{plaintop}
\restylefloat{table}

\usepackage[tableposition=top]{caption}

\newcommand{\bftab}{\fontseries{b}\selectfont}

\begin{document}

\title{An Extension of Fisher's Criterion: Theoretical Results with a Neural Network Realization}

\author{Ibrahim~Alsolami and Tomoki~Fukai \\ Okinawa Institute of Science and Technology (OIST)\\ Okinawa, Japan 904-0495}

\markboth{ }%
{Shell \MakeLowercase{\textit{et al.}}: Bare Demo of IEEEtran.cls for IEEE Journals}

\maketitle

\begin{abstract}

Fisher's criterion is a widely used tool in machine learning for feature selection. For large search spaces, Fisher’s criterion can provide a scalable solution to select features. A challenging limitation of Fisher’s criterion, however,  is that it performs poorly when mean values of class-conditional distributions are close to each other. Motivated by this challenge, we propose an extension of Fisher's criterion to overcome this limitation. The proposed extension utilizes the available heteroscedasticity of class-conditional distributions to distinguish one class from another. Additionally, we describe how our theoretical results can be casted into a neural network framework, and conduct a proof-of-concept experiment to demonstrate the viability of our approach to solve classification problems.

\end{abstract}


\IEEEpeerreviewmaketitle

\section*{Nomenclature}
\addcontentsline{toc}{section}{Nomenclature}
\begin{IEEEdescription}[\IEEEusemathlabelsep\IEEEsetlabelwidth{$V_1,V_2,V_3$}]
\item[$\displaystyle \vv{\bm{x}}$] A vector.
\item[$\displaystyle W$] A matrix.
\item[$\displaystyle w^{T}_{i}$]  $i^\text{th}$ row of matrix $W$.
\item[$\mathbb{R}$] The set of real numbers.

\item[$\displaystyle P(C_k)$] Class-prior probability.
\item[$\displaystyle P(x|C_k)$] Class-conditional distribution.
\item[$\displaystyle X \sim P$] Random variable $X$ has distribution $P$.

\item[$\displaystyle \mathcal{N}(x; \mu \,, \sigma^2)$]  Gaussian distribution over $x$ with mean $\mu$ and variance $\sigma^2$.

\end{IEEEdescription}

\section{Introduction}

\IEEEPARstart{F}{eature} selection plays a valuable role in machine learning. For a classification system, it identifies key features needed to distinguish between classes. With fewer features, a model is simplified, and its computational speed is likely to improve. Knowing which features to keep, however, is a challenging task. A dataset with $n$  features has $2^n$ possible combinations of features. As $n$ increases, the search space quickly becomes prohibitively large---rendering an exhaustive search approach impractical.

Fisher's criterion~\cite{Fisher} can provide a scalable solution to select features. In many applications, Fisher's criterion  is used to rank features by measuring their discriminative power. Once features are ranked, the best performing among them are selected for a model. Nowadays, Fisher's criterion is finding favor in a variety of applications, ranging from traffic-sign recognition~\cite{Zaklouta} to gene selection~\cite{Peng}.

Algebraically, Fisher’s criterion can be regarded as a generalized Rayleigh quotient~\cite{Duda}, which is used to quantify the separation between projected sets. From a regression perspective, Fisher’s criterion is a special case of least squares~\cite{Bishop}---that seeks to best fit a hyperplane between two classes of data. An implicit assumption made in Fisher’s criterion to construct such a hyperplane for classification is that data is homoscedastic (i.e., all classes have exactly the same variance/covariance matrix). This assumption simplifies the mathematical treatment. In practice, however, data is rarely homoscedastic. Heteroscedastic  (unequal variance/covariance matrix) data is largely expected. The interested reader is referred to~\cite{Duin,Zheng} for an insightful discussion.

The problem of classification in the presence of heteroscedasticity is  exacerbated when the mean values of class-conditional distributions are close to each other. When the mean values are far apart $|\mu_1- \mu_0|\gg 0$, the homoscedasticity assumption is reasonable because its impact on classification accuracy is practically negligible---since the overlap between class-conditional distributions is relatively small. However, as the mean values of class-conditional distributions approach each other, the small-overlap assumption becomes hard to justify. In this paper, we extend Fisher’s criterion to operate in this specific domain, where $\mu_0\approx \mu_1$.

In this study, a solution to the aforementioned difficulty is obtained  by formulating the problem as a mathematical expectation of a binary detection task. Through this formulation, an extension of Fisher’s criterion is derived that utilizes the heteroscedasticity of distributions to distinguish classes from each other.

\smallskip

\textbf{Contributions}. The main contributions of this work are threefold:
\smallskip

\begin{itemize}
  \setlength{\itemsep}{2pt}
  \setlength{\parskip}{0.9pt}
  \setlength{\parsep}{0pt}

\item We develop an extension of Fisher’s criterion to handle settings where mean values of class-conditional distribution are close to each other (Section~\ref{sec:tr}).

\item We describe how our theoretical results can be casted into  a neural network framework (Section~\ref{ps}).

\item We conduct a proof-of-concept experiment to demonstrate the viability of our approach (Section~\ref{er}).

\end{itemize}

\smallskip

The remainder of this paper is organized into five sections: In Section~\ref{sec:tr}, we derive the proposed extension of Fisher’s criterion. In Section~\ref{ps}, we present our neural network realization of our theoretical results. In Section~\ref{er}, we describe details of our experimental setup and discuss our experimental results.  In Section~\ref{sec:rw}, we outline related work.  In Section~\ref{sec:con}, we conclude the paper with a summary and identify future research directions.

\clearpage

\section{Proposed Extension} \label{sec:tr}

This section details the proposed extension of Fisher's criterion. We begin by listing our assumptions:

\vspace{.5cm}

\noindent \textbf{Assumption 1:}
\vspace{0.2cm}

\begin{equation}
P(x\,|\,C_k)=\mathcal{N}(x\,;\,\mu_k,\,\sigma_k^{2}), ~~~~~k\in \{0,1\}~.
\end{equation}

\vspace{0.2cm}

\noindent \textbf{Assumption 2:}

\begin{multline}\label{eq:def}
\hspace{-.4cm}\resizebox{1\hsize}{!}{$P(C_k)\,\mathbb{E}_{X\sim P(x|C_{k})} \Bigl[P(x|C_{k})\Bigr]>
P(C_{1\text{-}k})\,\mathbb{E}_{X\sim P(x|C_{k})} \Bigl[P(x|C_{1\text{-}k})\Bigr]$},\\
k\in\{0,1\}~;
\end{multline}

\noindent here  $P(C_k)$ is a class-prior probability, and $P(x|C_k)$ is a class-conditional distribution. The intuition behind  Assumption 2 is provided in Appendix~\ref{Assumption2}.

\bigskip

The mathematical expectation of  $P(x|C_{1\text{-}k})$  w.r.t.  $P(x|C_k)$  is

\begin{align}\label{eq:lnai}
\mathbb{E}_{X\sim P(x|C_{k})} \Bigl[P(x|C_{{1\text{-}k}})\Bigr]=\int_{-\infty}^{\infty} &P(x|C_{1\text{-}k}) P(x|C_k)  \,dx~, \nonumber\\
 & ~~~~~~~~~~~~~k\in\{0,1\}~;
\end{align}

\noindent here we  have a product of two Gaussian functions, $P(x|C_{1\text{-}k}) P(x|C_k)$; this product is a Gaussian function with:

\bigskip

\begin{align} 
&\text{a mean:~} \tilde{\mu}=\frac{\mu_0\sigma_1^2+\mu_1\sigma_0^2}{\sigma_1^2+\sigma_0^2}~, \hspace{1.5cm}  \label{eq:m}\\[1em]
&\text{standard deviation:~}\tilde{\sigma}= \frac{\sigma_0\,\sigma_1}{\sqrt{\sigma^2_1+\sigma^2_0}}~, \hspace{1.5cm}\label{eq:sd}\\[1em]
&\text{and scaling factor of:~} \tilde{S}=\frac{1}{\sqrt{2\pi (\sigma^2_1+\sigma^2_0)}}~~e^{-\frac{(\mu_1-\mu_0)^2}{2(\sigma_1^2+\sigma_{0}^2)}}~\hspace{1.5cm} \label{eq:sf}
\end{align}

\medskip

\noindent (see~\cite{Bromiley} for a detailed discussion). Using Eqs.~\ref{eq:m}--\ref{eq:sf} in \ref{eq:lnai}, yields

\medskip

\begin{align}
\mathbb{E}_{X\sim P(x|C_{k})} \Bigl[P(x|C_{{1\text{-}k}})\Bigr]=&~\tilde{S}\int_{-\infty}^{\infty} \mathcal{N}\left(x\,;\,\tilde{\mu}\,,\tilde{\sigma}^2\right)\,dx\nonumber\\
 =&~\frac{1}{\sqrt{2\pi (\sigma^2_1+\sigma^2_0)}}~~e^{-\frac{(\mu_1-\mu_0)^2}{2(\sigma_1^2+\sigma_{0}^2)}} \label{eq:E1}~~.
\end{align}

\medskip

\noindent Likewise,

\begin{align}
\mathbb{E}_{X\sim P(x|C_{k})} \Bigl[P(x|C_{k})\Bigr]=&~\int_{-\infty}^{\infty} P(x|C_k) P(x|C_k) \,dx \nonumber\\
 =&~\frac{1}{2\sqrt{\pi}\sigma_k},~~~k\in\{0,1\}\label{eq:E2}~~.
\end{align}

\vspace{1cm}
\noindent Substituting  Eqs. \ref{eq:E1} and  \ref{eq:E2} into  inequality~\ref{eq:def};  it follows that

\begin{equation}
     \label{eq:ine}
\mathcal{F}>  2\ln \left(\sqrt{\dfrac{2\sigma^2_k}{\sigma^2_1+\sigma^2_{0}}}\dfrac{P(C_{1\text{-}k)}}{P(C_{k})}\right),~~~k\in \{0,1\}~.
\end{equation}

\noindent here $\mathcal{F}\triangleq\frac{(\mu_1-\mu_0)^2}{\sigma^2_1+\sigma^2_0}$ is  Fisher’s criterion. Inequality~\ref{eq:ine} provides  a lower bound on Fisher’s criterion. Utilizing  this observation, the proposed  extension of Fisher’s criterion is

 \begin{equation} \label{eq:divergence}
\mathbf{\mathcal{D}}^k=\mathcal{F}-\mathcal{T}^k,~~~~k \in \{0,1\}~,
 \end{equation}

\noindent here  $\mathcal{T}^k$ is the RHS of inequality~\ref{eq:ine}. 
 The higher the value of $\mathbf{\mathcal{{D}}}^k$, the more discriminative power a feature holds.  It may be worthwhile to note that $\mathbf{\mathcal{{D}}}^k$ is capturing a divergence between $P(x|C_0)$ and $P(x|C_1)$. This divergence  depends not only on the difference between mean values, but also on a relative variance, enabling it to gauge the dissimilarity between distribution when $\mu_0 \approx \mu_1$\,$\Bigl(\text {or}~ \mathcal{F}=\frac{(\mu_1-\mu_0)^2}{\sigma^2_1+\sigma^2_0} \approx 0 \Bigr)$. In Section~\ref{er},  we will employ $\mathbf{\mathcal{{D}}}^k$ in our experiments to help select a subset of  features  used for image classification.

\section{Model} \label{ps}

In this section, we illustrate how our theoretical results in Section~\ref{sec:tr} can be casted into a neural network framework. We first propose a neural network architecture followed by a description of its activation functions. The design aim here is to provide a neural network realization of Eq.~\ref{eq:def} and show how Eq.~\ref{eq:divergence} can be employed in practice. In Section~\ref{er}, we will test the performance of the proposed network.

\subsection{Architecture and principles of operation}\label{sec:apo}

\begin{figure*}[tbh!]
\begin{center}
\includegraphics[scale=1]{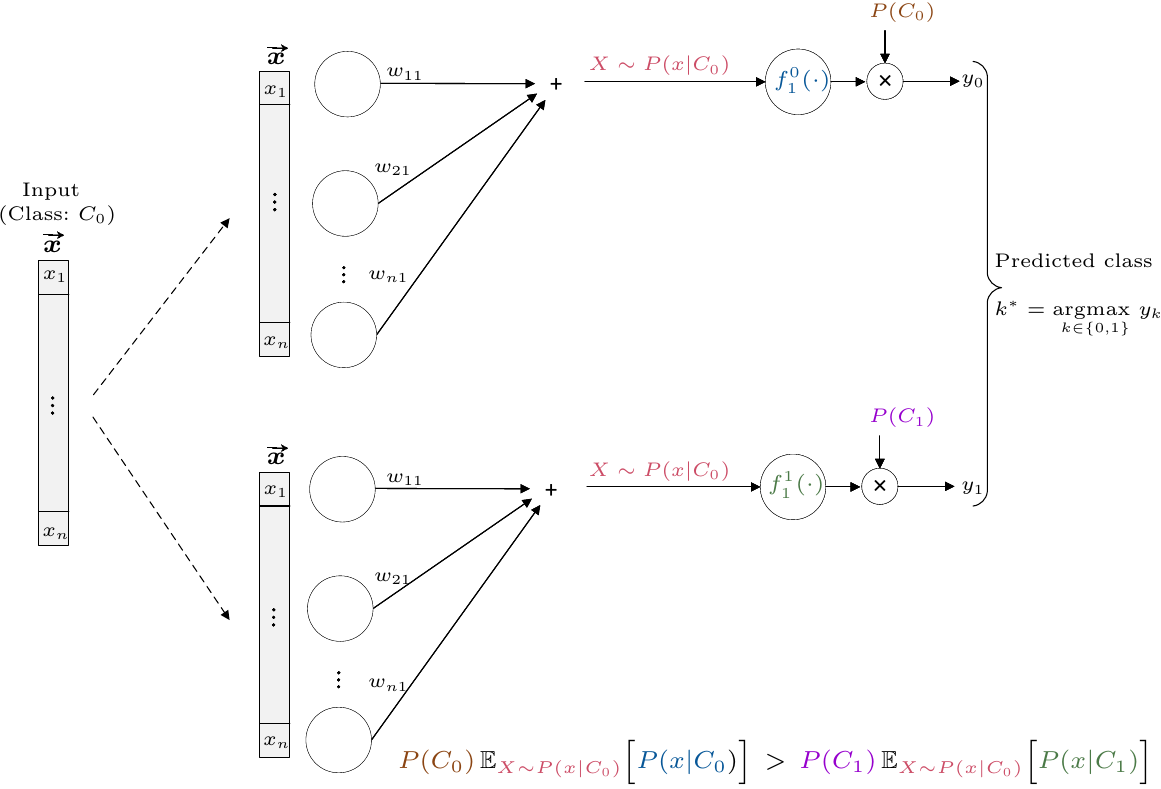} 
\end{center}
\caption{A  neural network realization of Eq.~\ref{eq:def}. The neural network and Eq.~\ref{eq:def} are both color-coded in this figure to better visualize the one-to-one correspondence between them. The top network is for the first class ($k=0$), while the bottom network is for the second class ($k=1$). Here, $f^k_i(\cdot)$  is an activation function (see Subsection~\ref{sec:saf} and Appendix~\ref{Assumption2}). The weights of both networks are identical. In this example, an input sample  belonging to class $C_0$ is presented to both networks, and each network outputs a corresponding  $y_k$ value. The class of the network that produces the maximum output  value, $y_k$, is declared the class of the input sample.}
\label{fig:CC}

\end{figure*}

Figure~\ref{fig:CC} is a schematic diagram to illustrate how Eq.~\ref{eq:def} can be realized. For simplicity,  the network in this illustration consists of a single node in the hidden layer for each class. Moreover, class-prior probabilities are realized by activation functions (Subsection \ref{sec:saf}).

\begin{figure*}%
    \centering
    \subfloat[]{\label{fig:PS}{\includegraphics[scale=0.845]{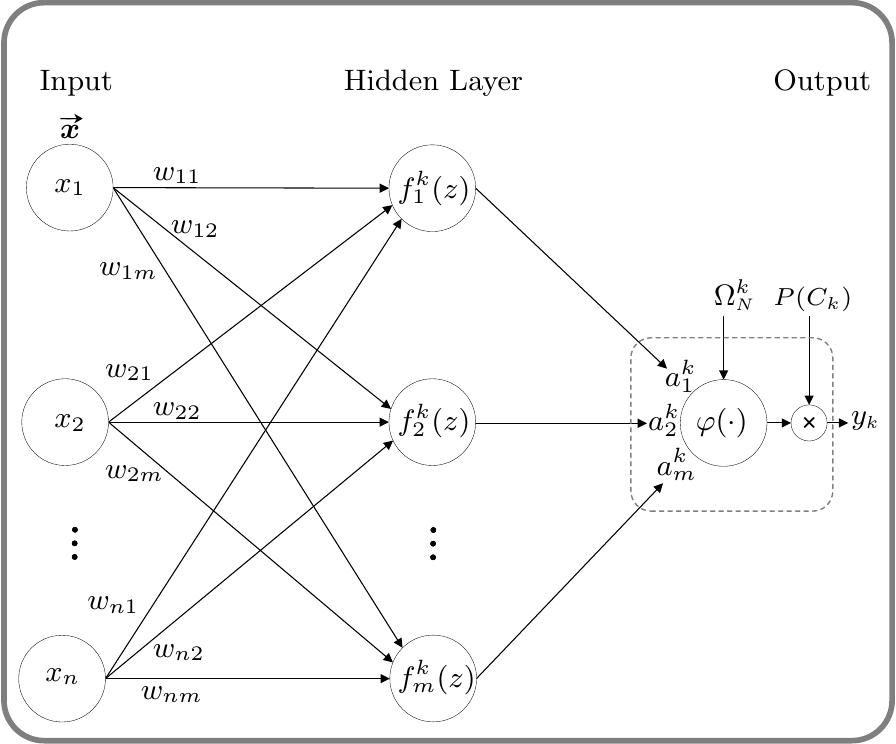} }}%
    \subfloat[]{\label{fig:AF}{\includegraphics[scale=0.845]{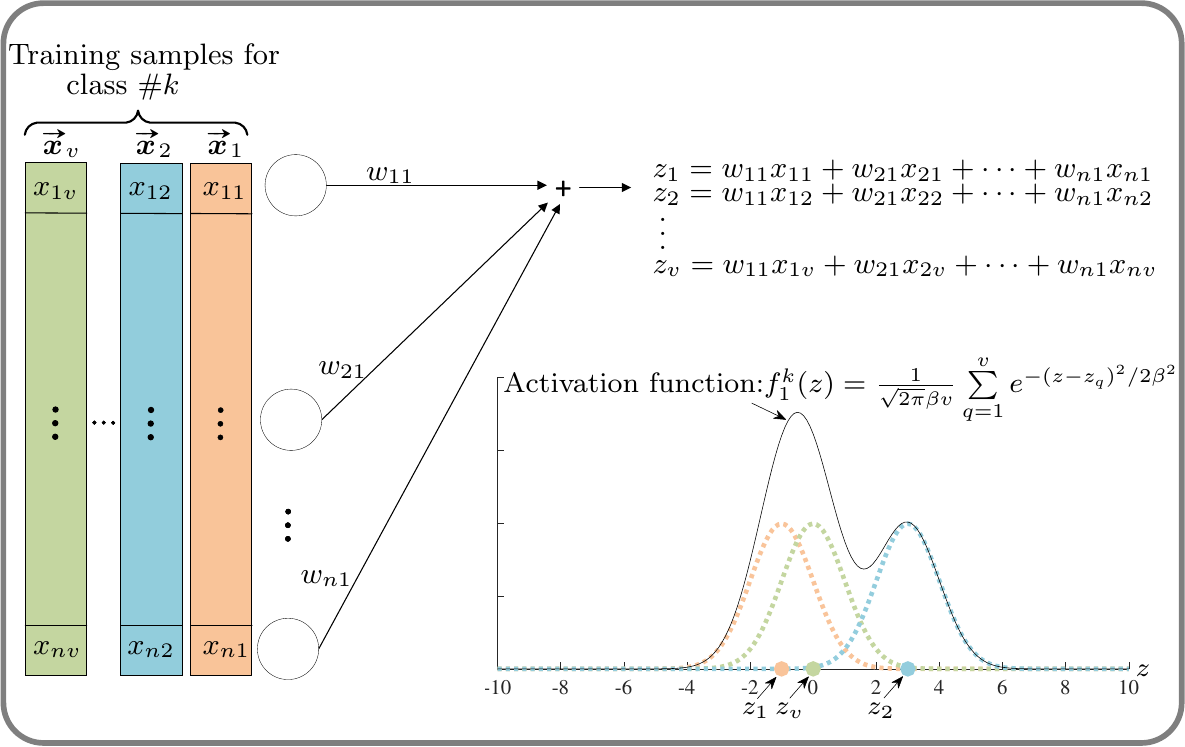} }} %
    \vspace{.56cm}
    \caption{(a) Architecture of proposed scheme.~~(b) Construction of activation functions.}%
    \label{fig:example}%
\end{figure*}

Figure~\ref{fig:PS} shows the general architecture of the proposed neural network, with $m$ hidden layers. The network consists of three layers: an input, hidden, and output layer. For each node in the hidden layer, an  input vector  $\vv{\bm{x}} \in \mathbb{R}^n$ is multiplied by random weights to form $z$, which is fed to  an activation function $f_i^k (z)$. Here $f_i^k (z)$ is the $i^{\text{th}}$ activation function for class $k$.

\begin{figure}[thb!]
\begin{center}
\includegraphics[scale=1.1]{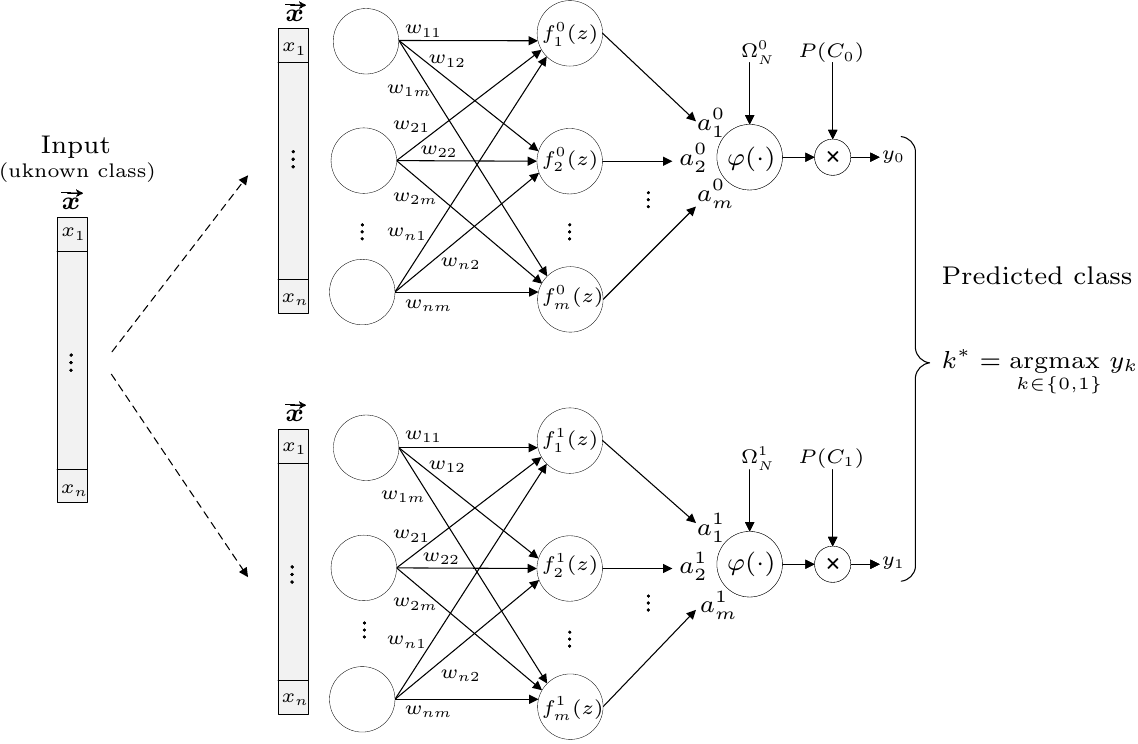}
\end{center}
\caption{Expanded illustration of proposed scheme. The top network is for the first class ($k=0$), while the bottom network is for the second class ($k=1$). The weights of both networks are identical. An input sample of an unknown class is presented to both networks, and each network outputs a corresponding  $y_k$ value. The class of the network that produces the maximum output value, $y_k$,  is declared the class of the input sample.}
\label{fig:PSA}
\end{figure}

Outputs of activation functions, represented by set $\mathcal{A}=\{a_1, a_2,\dots, a_m\}$, are in turn fed to $\varphi(\cdot)$. Function $\varphi(\cdot)$ selects (based on Eq.~\ref{eq:divergence}) a subset, $\Omega^{\scaleto{k}{4pt}}_{\scaleto{N}{3.3pt}}$ (Section~\ref{er}), of elements from $\mathcal{A}$ and sums them. This summation is computed for all classes ($k=0, 1$) and multiplied by corresponding class-prior probabilities, $P(C_k)$, to obtain a set of outputs values $\mathcal{Y}=\{y_0, y_1\}$. The predicted class is the argument of set $\mathcal{Y}$ that yields the maximum value: $\mbox{\footnotesize\( k^{*}=\aggregate{argmax}{k \in\{0,1\}}y_k \)}$. \textbf{Algorithm}~\ref{alg:PS} summarizes the proposed scheme, and an expanded illustration of the proposed scheme is provided in Fig.~\ref{fig:PSA} (Appendix~\ref{fig:Appendix3}).

\RestyleAlgo{ruled}

\SetKwComment{Comment}{$\triangleright$\ }{}

\begin{algorithm}[thb!]
    \DontPrintSemicolon
\caption{Proposed Scheme}\label{alg:PS}

  \KwInput{$\vv{\bm{x}}$, $\{P(C_k)\}$, $\{f_i^k (z)\}$, $\Omega^{\scaleto{k}{4pt}}_{\scaleto{N}{3.3pt}}$}

  \vspace{0.2cm}
  \KwOutput{$k^{*}$} 
  
  \vspace{-0.4cm}
  {\scriptsize{\texttt{\:\:\:\:~~~~~~~~~~~~~~~~~~~~~~$\triangleright$ Here $k^{*}$ is the predicted }}}\\
  
    {\scriptsize{\texttt{\:\:\:\:~~~~~~~~~~~~~~~~~~~~~~~ class of $\vv{\bm{x}}$.}}}
  
  \vspace{0.4cm}
$\begin{aligned}
W\gets\begin{bmatrix}[1.2]
    \horzbar & w^{T}_{1} & \horzbar \\
    \horzbar & w^{T}_{2} & \horzbar \\
             & \vdots    &          \\
    \horzbar & w^{T}_{m} & \horzbar
  \end{bmatrix}_{m \times n}
\end{aligned}$

\vspace{-1.5cm}

{\scriptsize{\texttt{\:\:\:\:~~~~~~~~~~~~~~~~~~~~~~$\triangleright$ \:Generate random matrix $W$.}}}\\

\vspace{1.5cm}

\For{$k=0,1$}{ 

 \For{$i=1,\dots,m$}{

$z_i\gets w^{T}_{i}\vv{\bm{x}}$
 \vspace{0.1cm}
 
 $a_i^{k} \gets f_i^k (z_i)$
 \vspace{0.3cm}
 }
  \vspace{0.1cm}

$y_k\gets P(C_k) \sum\limits_{j\in\Omega^{\scaleto{k}{3.5pt}}_{\scaleto{N}{3pt}}} a_j^{k} $
}

\vspace{- 1.2cm}
{\scriptsize{\texttt{\:\:\:\:~~~~~~~~~~~~~~~~~~~~~~~~~~$\triangleright$\: Function ~$\varphi$ in Fig.~\ref{fig:PS}.}}}

\vspace{1cm}

$k^{*}\gets \aggregate{argmax}{k\in\{0,\,1\}} y_k$

\end{algorithm}

\vspace{-0.3cm}

\subsection{Activation functions}\label{sec:saf}

The key idea here is to employ class-conditional probability density functions, obtained from training samples, as activation functions,  $f_i^k(x)$. In Section~\ref{sec:tr} we have $P(x|C_k)$, and a way to realize such density functions in practice is to represent them as activation functions---that is, we use $f_i^k(x)$ to mimic $P(x|C_k)$. Fig.~\ref{fig:AF} illustrates how an activation function is constructed.  Each class, $k$, has a dedicated network (Fig.~\ref{fig:PS}). For each node in this network, training data $\{\vv{\bm{x}}_1,\vv{\bm{x}}_2,\dots,\vv{\bm{x}}_v\}$, of the same class, are multiplied by random weights to obtain a sample set  $\{z_1,z_2,\dots, z_v\}$. This set is used by a kernel density estimator (KDE) to construct an activation function as follows:

\begin{equation}
f_i^k (z)=\frac{1}{\sqrt{2 \pi}\beta v}\sum\limits_{q=1}^{v} e^{-(z-z_q)^2/2\beta^2},
\end{equation}

\noindent here subscript $i$ indicates the $i^{\text{th}}$ node of a given class $k$, while $v$ and $\beta$ denote the number of training samples and bandwidth of the KDE, respectively. A variety of methods can be used to find an apt value of $\beta$ (see, for instance,~\cite{Silverman,Sheather,Scott}, as well as \cite{Heidenreich} for a review). In our experiment, functions $f_i^k(z)$ are computed on-demand rather than  saved as lookup tables; this approach is computationally heavy but saves a great amount of memory.  The proposed neural network learns without weight tuning: learning is accomplished by letting data of training samples shape activation functions (Fig.~\ref{fig:AF}). \textbf{Algorithm}~\ref{alg:AF} provides a summary of how activation functions are constructed.

\begin{algorithm}[h!]
    \DontPrintSemicolon
\caption{Construction of Activation Functions}\label{alg:AF}
\vspace{0.3cm}
  \KwInput{$ \{\mathbf{x}^k_q\}^1_{k=0}$} 
  
    \vspace{-0.55cm}

  {\scriptsize{\texttt{~~~~~~~~~~~~~~~~~~~$\triangleright$ Training set; $\mathbf{x}^k_q \in\mathbb{R}^n $, where $k$ }}}
  
 {\scriptsize{\texttt{~~~~~~~~~~~~~~~~~~~~~indicates a given class, and $q$ is}}}
 
  {\scriptsize{\texttt{~~~~~~~~~~~~~~~~~~~~~is a sample's index.}}}

  \vspace{.5cm}
  \KwOutput{$\{f_i^k (z)\}$} 
  
   \vspace{-.5cm}
    {\scriptsize{\texttt{~~~~~~~~~~~~~~~~~~~$\triangleright$ Here $\{f_i^k (z)\}$ is the set of}}}
    
       {\scriptsize{\texttt{~~~~~~~~~~~~~~~~~~~~~activation function.}}}
  
  \vspace{0.4cm}
$\begin{aligned}
W\gets\begin{bmatrix}[1.2]
    \horzbar & w^{T}_{1} & \horzbar \\
    \horzbar & w^{T}_{2} & \horzbar \\
             & \vdots    &          \\
    \horzbar & w^{T}_{m} & \horzbar
  \end{bmatrix}_{m \times n} 
\end{aligned}$

\vspace{-2cm}

{\scriptsize{\texttt{~~~~~~~~~~~~~~~~~~~~~~~~~$\triangleright$ Generate random matrix $W$. }}}

{\scriptsize{\texttt{~~~~~~~~~~~~~~~~~~~~~~~~~~ This matrix  is identical  }}}

{\scriptsize{\texttt{~~~~~~~~~~~~~~~~~~~~~~~~~~  to $W$ of Algorithm~\ref{alg:PS}.}}}

\vspace{1.5cm}

\For{$k=0,1$}{ 

 \For{$i=1,\dots,m$}{
  \For{$q=1,\dots,v_{\scaleto{k}{3.5pt}}$}{

\vspace{.3cm}
$z_q\gets w^{T}_{i} \mathbf{x}^k_q$
\vspace{0.2cm}

 }

 \vspace{0.4cm}

 $\beta_i\gets  {\footnotesize	{\text{ Compute bandwidth from set } \{z_1,z_2,\dots,z_{v_{\scaleto{k}{2.5pt}}}\}}}$
 
  \vspace{.2cm}
 
  $f_i^k(z)\gets\frac{1}{\sqrt{2 \pi}\beta_i v_{\scaleto{k}{2.2pt}}}\sum\limits_{q=1}^{v_{\scaleto{k}{2.5pt}}} e^{-(z-z_q)^2/2\beta_i^2}$

 }
} 
\end{algorithm}

\vspace{5cm}

\section{Experiment and Results} \label{er}

\noindent \textbf{Task description:} Binary classification.

\bigskip

\noindent \textbf{Dataset:} Pairs of image classes are obtained from the MNIST~\cite{MNIST} and Fashion-MNIST dataset~\cite{FMNIST}; both datasets were $z$-normalized.

\begin{figure*}[thb!]
\begin{center}
\includegraphics[scale=1]{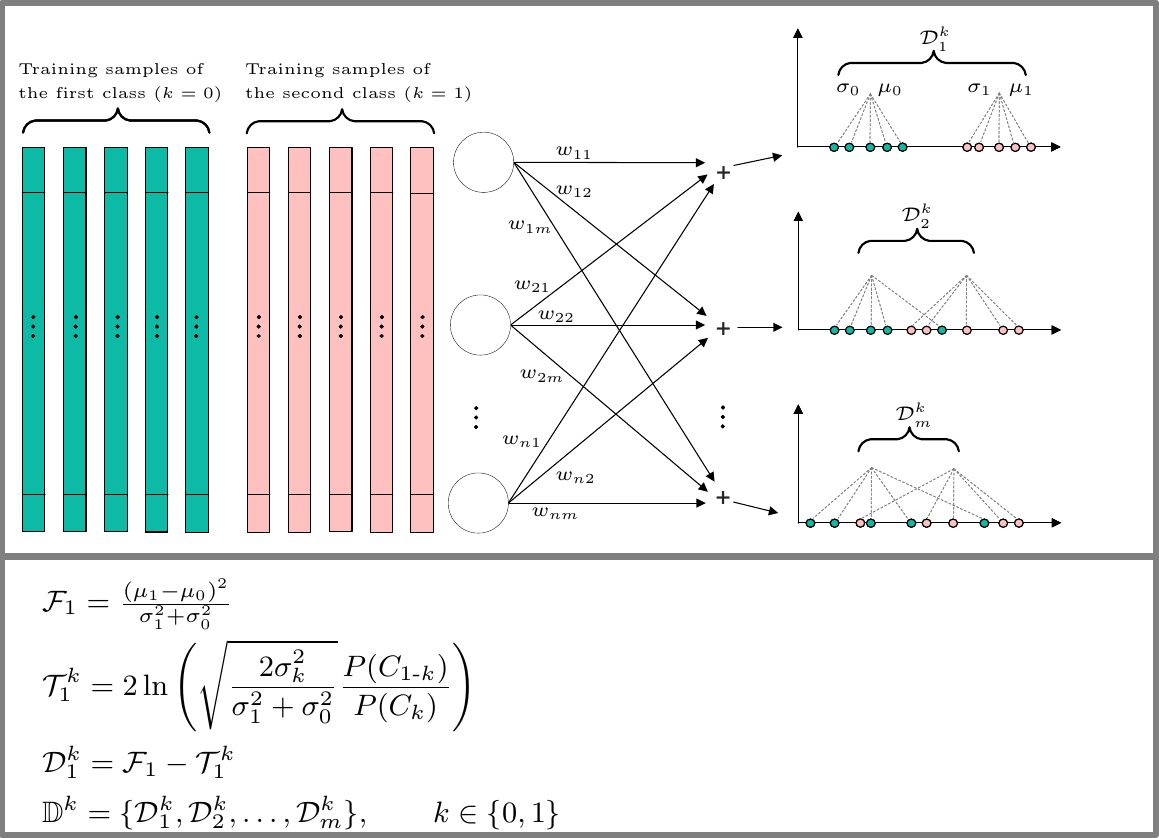}
\end{center}
\caption{Illustration for the construction of set $\mathbb{D}^k$, where superscript $k$ labels a class. Consider the uppermost node in the network for which we need to compute $\mathbf{\mathcal{{D}}}^k_1$. Here samples of each class are multiplied by random weights $ w^{T}_{1}=[w_{11},w_{21},\dots, w_{n1}]$. After which, the values of $\mu_i$ and $\sigma_i$ are computed, $i\in\{0,1\}$ (Section~\ref{er}). Using $\mu_i$ and $\sigma_i$ with  $P(C_i)$, the value of the divergence at the node under consideration is  $\mathbf{\mathcal{{D}}}^k_1=\mathcal{F}_1-\mathcal{T}^k_1$. This process is repeated for all nodes in the network to obtain  set $\mathbb{D}^k=\{ \mathbf{\mathcal{{D}}}^k_1, \mathbf{\mathcal{{D}}}^k_2,\dots, \mathbf{\mathcal{{D}}}^k_m \}$.}
\label{fig:Ds}
\end{figure*}

\bigskip
\noindent \textbf{Methods:}
\begin{itemize}

\item \textbf{Kernel density estimator (KDE).} 
Gaussian functions, $K(u)=\frac{1}{\sqrt{2 \pi}} e^{-\frac{1}{2}u^2}$, are used as  kernels  for the KDE, $f(z)=\frac{1}{\beta v}\sum\limits_{q=1}^{v} K\left(\frac{z-z_q}{\beta}\right)$. The bandwidth of  kernels are calculated  as $\beta=  \left(\frac{4}{3v}\right)^{\frac{1}{5}}\sigma$, where $\sigma$ is a measure of dispersion (spread) and $v$ is the number of samples of a set~\cite{Silverman}. The standard deviation can be used to  measure dispersion. In our experiment, however, we chose the median absolute deviation, $\eta$, as it provides a more robust dispersion measure when outliers are present $\left(\sigma=\frac{\eta}{0.6745},~\text{\cite{Viertl,Forstner}}\right)$.

\bigskip

\item \textbf{Weights}. The traditional approach to train neural networks is weight tuning, whereby weights of a network are computed either iteratively or by solving a set of  equations so as to minimize a discrepancy between a model’s prediction and a corresponding ground truth~\cite{RBF,ELM,Schmidhuber2}.  In this study, we provide an alternative approach. We illustrate how neural networks can be designed to classify patterns without weight tuning: weights of networks herein are randomly generated and left afterward unchanged. The  weights, $w_{ij}$, of the network (Fig.~\ref{fig:PS}) are drawn independently from a standard normal distribution, $w_{ij}\sim \mathcal{N}(0\,,\,1)$. 

\bigskip

\item \textbf{Difference between mean values  $\Delta \mu$}. After input vectors are multiplied by weights, the difference between mean values of distributions $\left(\Delta \mu=\mu_1-\mu_0\right)$ is adjusted to simulate a condition where $\mu_0 \approx \mu_1$. This is accomplished by letting $\Delta\mu$ draw values from a standard normal distribution, $\Delta\mu\sim \mathcal{N}(0\,,\,1)$.

\bigskip

\item \textbf{Divergence calculations}. To deal with outliers when calculating  the divergence in Eq.~\ref{eq:divergence}, we use the interquartile mean to approximate $\mu$~\cite{Sprent}, and  $\sigma\approx\frac{\eta}{0.6745}$ to approximate the standard deviation. Priori probabilities, $P(C_k)$, are estimated as the proportion of each class in a training dataset.

\bigskip

\item \textbf{Subset $\Omega^{\scaleto{k}{4pt}}_{\scaleto{N}{3.3pt}}$}. Consider Fig.~\ref{fig:Ds} (Appendix~\ref{fig:Appendix3}). $\Omega^{\scaleto{k}{4pt}}_{\scaleto{N}{3.3pt}}$ is a subset of the nodes that have the largest $N$ divergence values.  Let $\mathbb{D}
^k=\{ \mathbf{\mathcal{{D}}}^k_1,\, \mathbf{\mathcal{{D}}}^k_2,\,\dots,\, \mathbf{\mathcal{{D}}}^k_m \}$ denote a set of divergence values, where $ \mathbf{\mathcal{{D}}}^k_i$ denotes the divergence for class $k$  measured at the $ i^\text{th}$ node, and  $m$ is the total number of nodes in the hidden layer. The value of a given $ \mathbf{\mathcal{{D}}}^k_i$ is computed via Eq.~\ref {eq:divergence}. Additionally, let $\Omega^{\scaleto{k}{4pt}}=\{\omega_1,\dots,\omega_{\scaleto{N}{3.3pt}},\dots,\omega_{\scaleto{m}{3.3pt}}\}$ denote the set of indices of the sorted values of $\mathbb{D}^k$ in descending order. For example, if $\mathbb{D}^k=\{\mathbf{\mathcal{{D}}}^k_1,\mathbf{\mathcal{{D}}}^k_2,\mathbf{\mathcal{{D}}}^k_3\}$ and $\mathbf{\mathcal{{D}}}^k_2> \mathbf{\mathcal{{D}}}^k_3>\mathbf{\mathcal{{D}}}^k_1$, then $\Omega^{\scaleto{k}{4pt}}=\{2,3,1\}$.
 To keep the notation light,  subset $\Omega^{\scaleto{k}{4pt}}_{\scaleto{N}{3.3pt}}=\{\omega_1,\omega_2,\dots,\omega_{\scaleto{N}{3.3pt}}\}$  is used to denote the first $N$ values in set $\Omega^{\scaleto{k}{4pt}}$. In our experiments, $N=10$, and the total number of nodes in the hidden layer of the network is $m=10^4$.

\bigskip

\item \textbf{Randomization.} We run our network with 10 different randomization seeds to report a 95$\%$ confidence interval. The randomization is in the weights of the network.

\end{itemize}

\bigskip

\noindent \textbf{Experimental results and discussion:} The proposed extension has a potential to improve performance. Tables~\ref{tab:MNIST} and ~\ref{tab:FMNIST} present the classification accuracy of the proposed extension versus Fisher’s criterion.  On average, the proposed extension increases the classification accuracy of Fisher’s criterion by 25.50$\%$ for the MNIST dataset and  27.97$\%$ for the Fashion-MNIST dataset. This gain can be, in part, attributed to the ability of the proposed extension to utilize the available heteroscedasticity of class-conditional distributions. For example, if $\mu_0\approx\mu_1$, then the proposed extension would rank features that have a small variance higher than those with large ones (Eq.~\ref{eq:divergence}).

\begin{figure*}[tbh!]

\centering
\begin{tabular}{ccccc}

&
\subfloat[Samples from the Fashion-MNIST dataset.]{\includegraphics[scale=0.9]{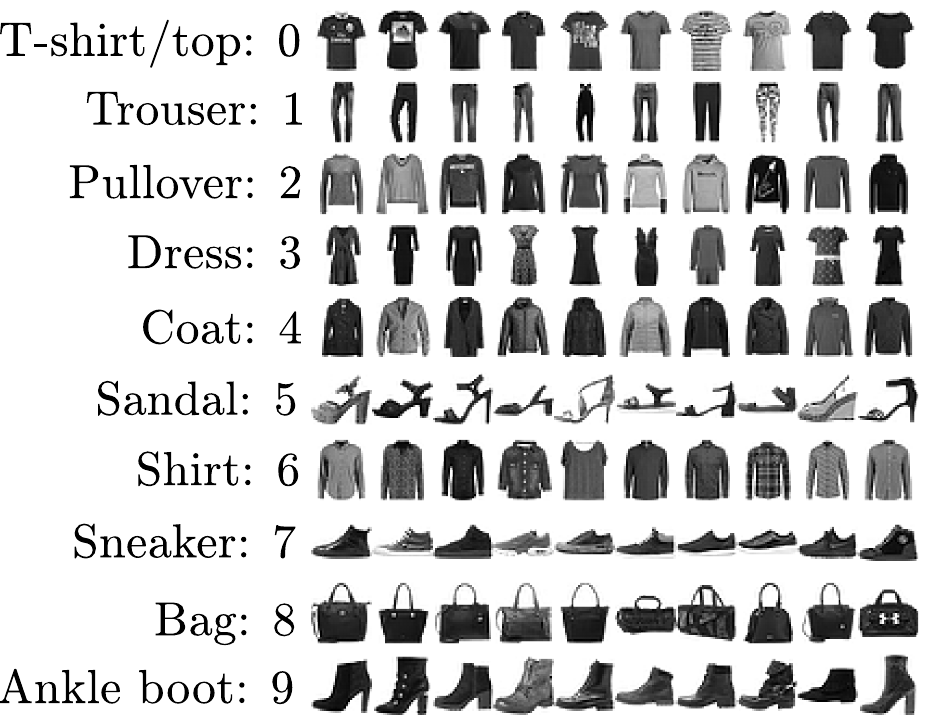} }&
\subfloat[Samples from the MNIST dataset.]{\includegraphics[scale=0.9]{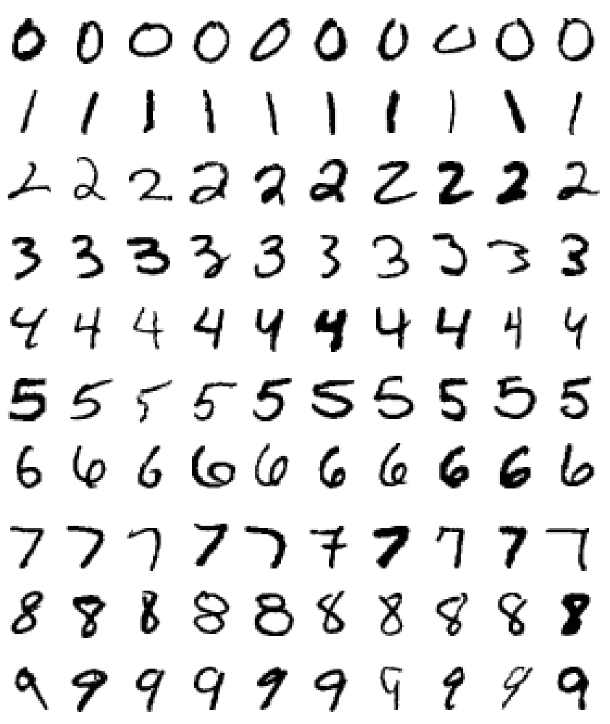}}&
\\

\vspace{0.75cm}
\\

 &
\subfloat[]{\includegraphics[scale=0.37]{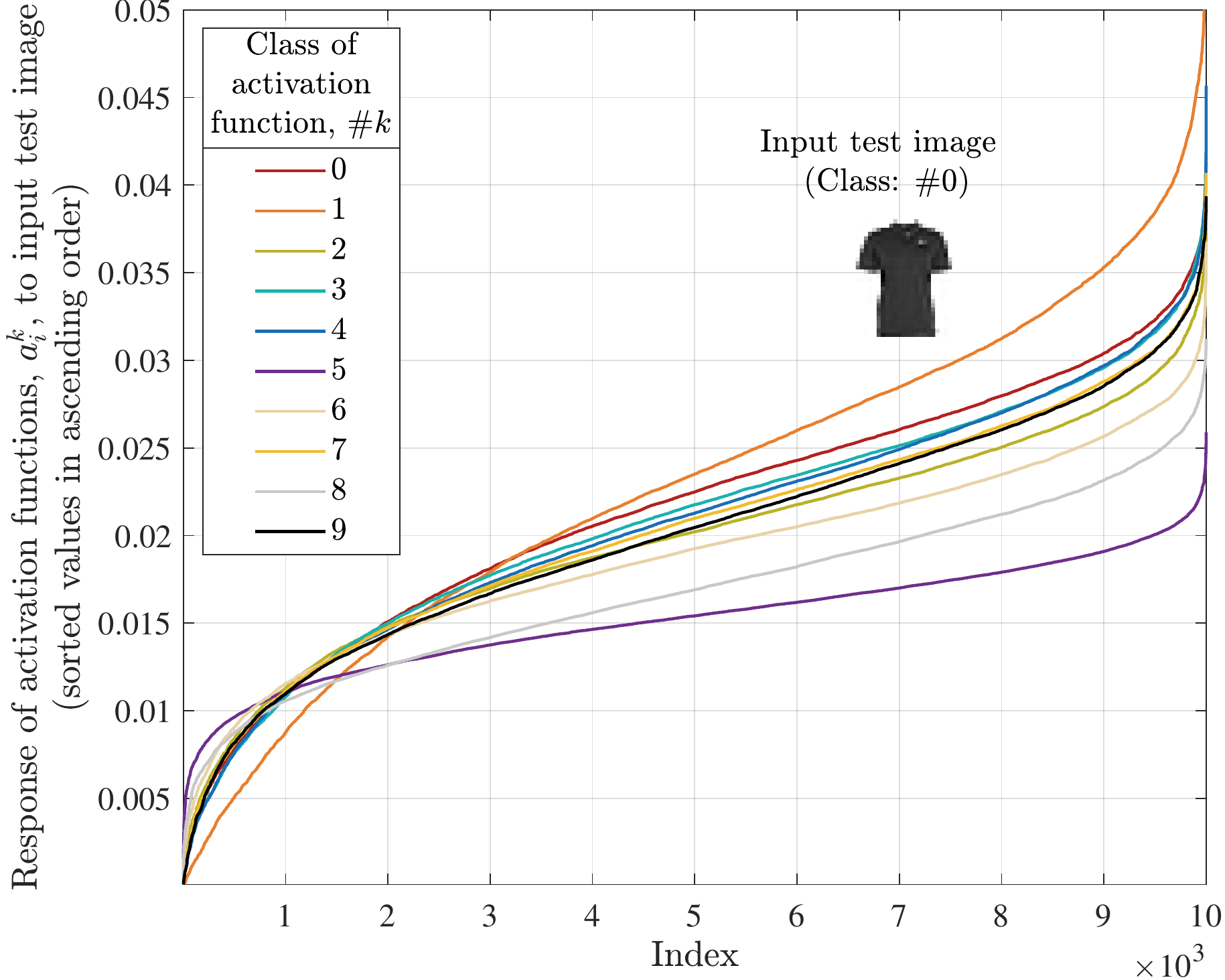} }&
\subfloat[]{\includegraphics[scale=0.37]{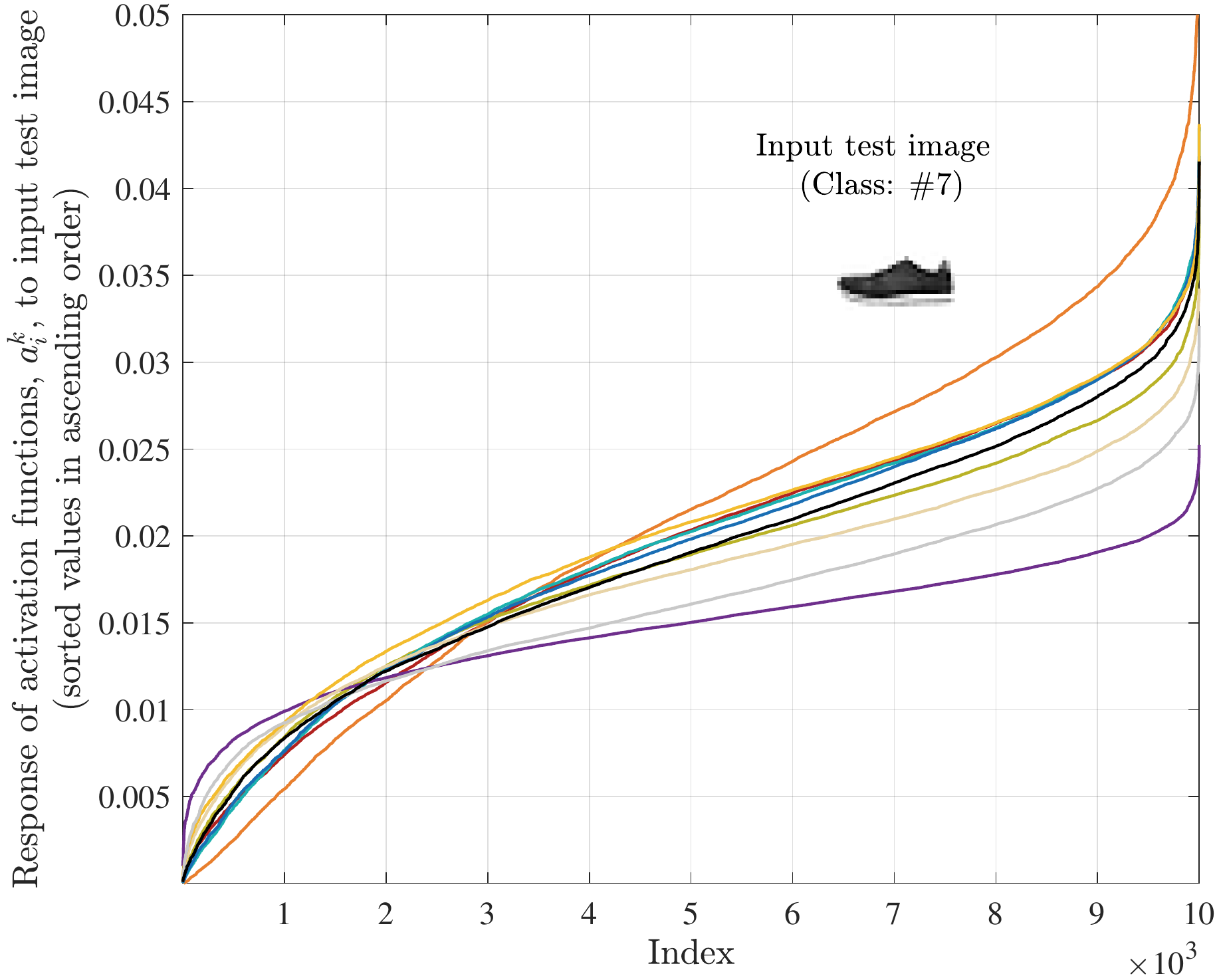}}&
\\
 &
\subfloat[]{\includegraphics[scale=0.37]{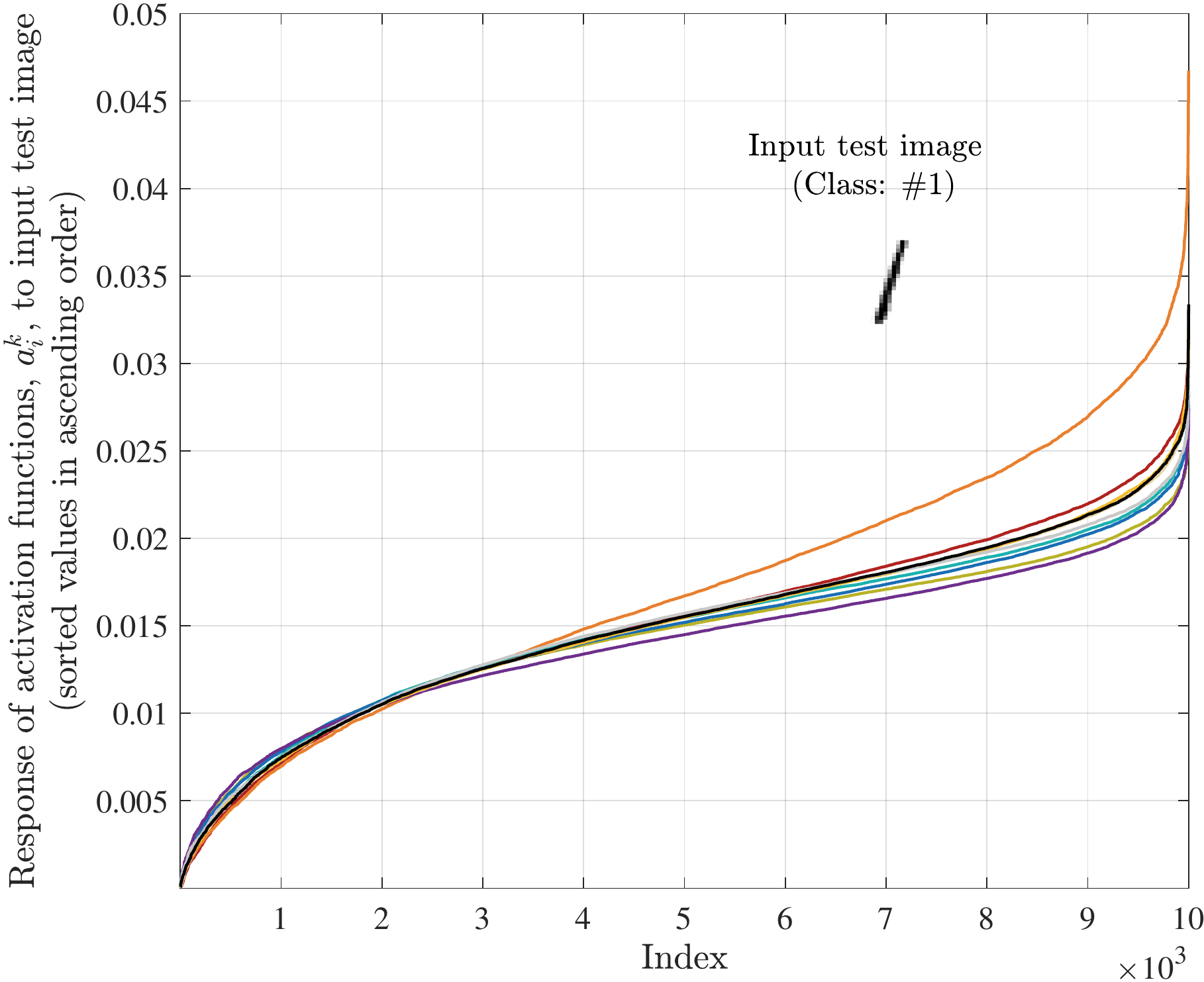} }&
\subfloat[]{\includegraphics[scale=0.37]{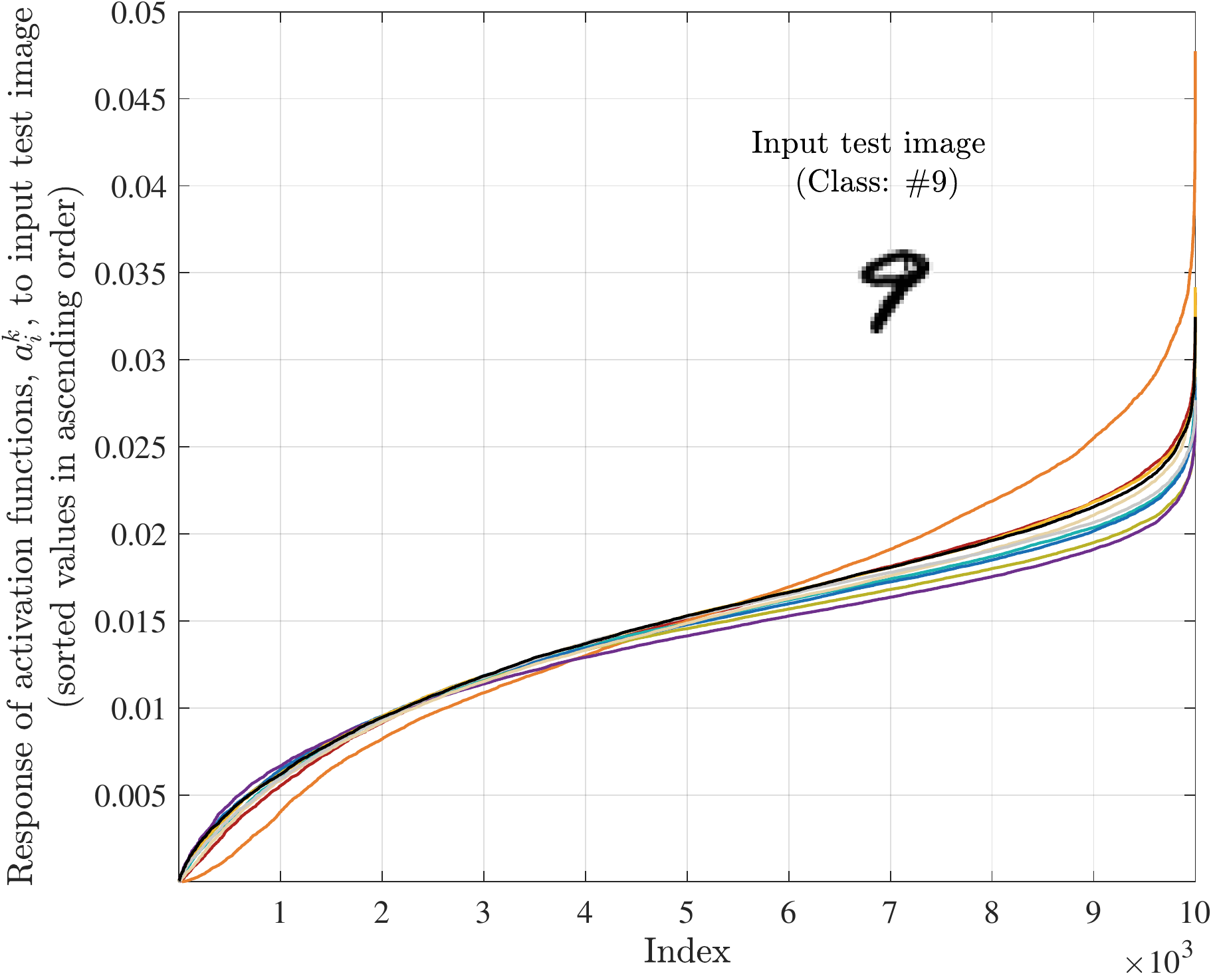}}&

\end{tabular}
\caption{(c)--(f) Responses of activation functions, $a^k_i$, to unseen test images.  The $y$-axis displays the sorted values of set $\{a^k_1,a^k_2,\dots,a^k_{m}\}$ in ascending order, while $x$-axis is the index of said sorted values (see Fig.~\ref{fig:PS} for a schematic illustration depicting $a^k_i$ responses). The number of nodes in the hidden layer of the network is $m=10^4$.}
\label{fig:sorted}
\end{figure*}

A reason for the proposed extension to perform slightly better in the Fashion-MNIST dataset than in the MNIST dataset, is that responses of activation functions are more spread out for the Fashion-MNIST dataset than in the MNIST dataset. This observation is displayed Figs.~\ref{fig:sorted}\,(c)--(f).  The further apart responses of activation functions are from each other, the less the chance of mistaking one class for another.



\setlength\tabcolsep{6pt}
\begin{table*}[tbh!]
   \caption{Classification Accuracy$^\dagger$ \\ {\scriptsize  MNIST Dataset}\\ {\scriptsize Difference between mean values:  $\Delta\mu\sim \mathcal{N}(0,1)$}   \\{\scriptsize \textcolor{black}{$\bullet $\, Ours\,(black)}~~~\textcolor{mygray}{ $\bullet$\,Fisher\,(gray)}}}  
   \label{tab:MNIST}
   \tiny
   \centering
   \begin{tabular}{c|cccccccccc}
   \toprule\toprule
  
   \scriptsize \textbf{Class $\#$} &  \scriptsize 0 &  \scriptsize  1&  \scriptsize  2&  \scriptsize  3 &  \scriptsize  4 &  \scriptsize 5 & \scriptsize  6 &  \scriptsize 7 &  \scriptsize  8 &  \scriptsize  9 \\ 
     \hline

   \multirow{2}{*}{ \scriptsize 0} & - & \textcolor{black}{\bftab92.89\%$\pm$0.55\%}& \textcolor{black}{\bftab84.87\%$\pm$0.93\%} & \textcolor{black}{\bftab86.23\%$\pm$1.15\%}& \textcolor{black}{\bftab90.74\%$\pm$0.84\%} &   \textcolor{black}{\bftab86.55\%$\pm$1.06\%} & \textcolor{black}{\bftab87.68\%$\pm$0.82\%}  & \textcolor{black}{\bftab92.91\%$\pm$0.85\%} & \textcolor{black}{\bftab85.45\%$\pm$0.75\%} &\textcolor{black}{\bftab91.87\%$\pm$0.65\%}\\
                                       &  & \textcolor{mygray}{\bftab62.43\%$\pm$1.95\%} & \textcolor{mygray}{\bftab57.70\%$\pm$1.69\%} & \textcolor{mygray}{\bftab59.00\%$\pm$2.16\%}  & \textcolor{mygray}{\bftab59.89\%$\pm$1.52\%}  &\textcolor{mygray}{\bftab59.36\%$\pm$1.23\%}  & \textcolor{mygray}{\bftab63.33\%$\pm$1.87\%} & \textcolor{mygray}{\bftab62.88\%$\pm$0.82\%}  & \textcolor{mygray}{\bftab60.44\%$\pm$1.42\%} & \textcolor{mygray}{\bftab64.12\%$\pm$1.35\%} \\
    \hline   
    
     \multirow{2}{*}{ \scriptsize 1} & \textcolor{black}{\bftab92.89\%$\pm$0.55\%} & - & \textcolor{black}{\bftab83.44\%$\pm$1.28\%} & \textcolor{black}{\bftab82.81\%$\pm$1.39\%}& \textcolor{black}{\bftab83.00\%$\pm$1.33\%} &   \textcolor{black}{\bftab90.40\%$\pm$0.98\%} & \textcolor{black}{\bftab90.37\%$\pm$1.66\%}  & \textcolor{black}{\bftab89.95\%$\pm$1.10\%} & \textcolor{black}{\bftab82.46\%$\pm$1.60\%} &\textcolor{black}{\bftab87.98\%$\pm$1.77\%}\\
                                       & \textcolor{mygray}{\bftab62.43\%$\pm$1.95\%} &  & \textcolor{mygray}{\bftab58.84\%$\pm$0.43\%} & \textcolor{mygray}{\bftab60.54\%$\pm$1.18\%}  & \textcolor{mygray}{\bftab60.06\%$\pm$0.85\%}  &\textcolor{mygray}{\bftab61.99\%$\pm$0.71\%}  & \textcolor{mygray}{\bftab63.05\%$\pm$1.67\%} & \textcolor{mygray}{\bftab61.14\%$\pm$1.75\%}  & \textcolor{mygray}{\bftab60.37\%$\pm$0.61\%} & \textcolor{mygray}{\bftab61.68\%$\pm$2.08\%} \\
    \hline

 \multirow{2}{*}{\scriptsize 2} & \textcolor{black}{\bftab84.87\%$\pm$0.93\%} & \textcolor{black}{\bftab83.44\%$\pm$1.28\%}& -& \textcolor{black}{\bftab84.00\%$\pm$0.96\%}& \textcolor{black}{\bftab87.72\%$\pm$1.19\%} &   \textcolor{black}{\bftab87.28\%$\pm$1.25\%} & \textcolor{black}{\bftab87.33\%$\pm$1.33\%}  & \textcolor{black}{\bftab86.57\%$\pm$1.82\%} & \textcolor{black}{\bftab83.78\%$\pm$1.51\%} &\textcolor{black}{\bftab86.82\%$\pm$1.91\%}\\
                                       & \textcolor{mygray}{\bftab57.70\%$\pm$1.69\%} & \textcolor{mygray}{\bftab58.84\%$\pm$0.43\%} &   & \textcolor{mygray}{\bftab60.02\%$\pm$1.34\%}  & \textcolor{mygray}{\bftab60.29\%$\pm$2.63\%}  &\textcolor{mygray}{\bftab61.52\%$\pm$1.50\%}  & \textcolor{mygray}{\bftab59.17\%$\pm$1.62\%} & \textcolor{mygray}{\bftab60.18\%$\pm$1.49\%}  & \textcolor{mygray}{\bftab59.57\%$\pm$2.13\%} & \textcolor{mygray}{\bftab59.79\%$\pm$2.05\%} \\
    \hline

 \multirow{2}{*}{ \scriptsize 3} & \textcolor{black}{\bftab86.23\%$\pm$1.15\%} & \textcolor{black}{\bftab82.81\%$\pm$1.39\%}& \textcolor{black}{\bftab84.00\%$\pm$0.96\%} & - & \textcolor{black}{\bftab88.56\%$\pm$1.14\%} &   \textcolor{black}{\bftab82.82\%$\pm$1.16\%} & \textcolor{black}{\bftab88.48\%$\pm$1.14\%}  & \textcolor{black}{\bftab89.03\%$\pm$1.54\%} & \textcolor{black}{\bftab82.62\%$\pm$1.43\%} &\textcolor{black}{\bftab87.15\%$\pm$1.23\%}\\
                                       & \textcolor{mygray}{\bftab59.00\%$\pm$2.16\%} & \textcolor{mygray}{\bftab60.54\%$\pm$1.18\%} & \textcolor{mygray}{\bftab60.01\%$\pm$1.34\%} &   & \textcolor{mygray}{\bftab60.43\%$\pm$1.41\%}  &\textcolor{mygray}{\bftab61.27\%$\pm$0.84\%}  & \textcolor{mygray}{\bftab62.33\%$\pm$1.45\%} & \textcolor{mygray}{\bftab62.99\%$\pm$1.82\%}  & \textcolor{mygray}{\bftab61.13\%$\pm$0.83\%} & \textcolor{mygray}{\bftab62.00\%$\pm$1.37\%} \\
    \hline

     \multirow{2}{*}{ \scriptsize 4} & \textcolor{black}{\bftab90.74\%$\pm$0.84\%} & \textcolor{black}{\bftab83.00\%$\pm$1.33\%}& \textcolor{black}{\bftab87.72\%$\pm$1.19\%} & \textcolor{black}{\bftab88.56\%$\pm$1.14\%}& - &   \textcolor{black}{\bftab88.68\%$\pm$0.84\%} & \textcolor{black}{\bftab88.59\%$\pm$0.47\%}  & \textcolor{black}{\bftab83.94\%$\pm$1.67\%} & \textcolor{black}{\bftab84.87\%$\pm$0.96\%} &\textcolor{black}{\bftab76.98\%$\pm$1.61\%}\\
                                       & \textcolor{mygray}{\bftab59.89\%$\pm$1.52\%} & \textcolor{mygray}{\bftab60.06\%$\pm$0.85\%} & \textcolor{mygray}{\bftab60.29\%$\pm$2.63\%} & \textcolor{mygray}{\bftab60.44\%$\pm$1.41\%}  &   &\textcolor{mygray}{\bftab61.07\%$\pm$1.76\%}  & \textcolor{mygray}{\bftab59.89\%$\pm$1.94\%} & \textcolor{mygray}{\bftab60.16\%$\pm$1.62\%}  & \textcolor{mygray}{\bftab60.85\%$\pm$1.47\%} & \textcolor{mygray}{\bftab58.87\%$\pm$1.06\%} \\
    \hline

     \multirow{2}{*}{ \scriptsize 5} & \textcolor{black}{\bftab86.55\%$\pm$1.06\%} & \textcolor{black}{\bftab90.40\%$\pm$0.98\%}& \textcolor{black}{\bftab87.28\%$\pm$1.25\%} & \textcolor{black}{\bftab82.82\%$\pm$1.16\%}& \textcolor{black}{\bftab88.68\%$\pm$0.84\%} &  - & \textcolor{black}{\bftab87.21\%$\pm$1.75\%}  & \textcolor{black}{\bftab89.69\%$\pm$0.69\%} & \textcolor{black}{\bftab79.31\%$\pm$1.43\%} &\textcolor{black}{\bftab85.83\%$\pm$1.58\%}\\
                                       & \textcolor{mygray}{\bftab59.36\%$\pm$1.23\%} & \textcolor{mygray}{\bftab61.99\%$\pm$0.71\%} & \textcolor{mygray}{\bftab61.52\%$\pm$1.50\%} & \textcolor{mygray}{\bftab61.27\%$\pm$0.84\%}  & \textcolor{mygray}{\bftab61.07\%$\pm$1.76\%}  &   & \textcolor{mygray}{\bftab59.58\%$\pm$1.52\%} & \textcolor{mygray}{\bftab61.39\%$\pm$1.79\%}  & \textcolor{mygray}{\bftab59.90\%$\pm$1.41\%} & \textcolor{mygray}{\bftab60.32\%$\pm$1.50\%} \\
    \hline

     \multirow{2}{*}{ \scriptsize 6} & \textcolor{black}{\bftab87.68\%$\pm$0.82\%} & \textcolor{black}{\bftab90.37\%$\pm$1.66\%}& \textcolor{black}{\bftab87.33\%$\pm$1.33\%} & \textcolor{black}{\bftab88.48\%$\pm$1.14\%}& \textcolor{black}{\bftab88.59\%$\pm$0.47\%} &   \textcolor{black}{\bftab87.21\%$\pm$1.75\%} & -  & \textcolor{black}{\bftab91.84\%$\pm$0.68\%} & \textcolor{black}{\bftab84.14\%$\pm$1.26\%} &\textcolor{black}{\bftab90.52\%$\pm$1.15\%}\\
                                       & \textcolor{mygray}{\bftab63.33\%$\pm$1.87\%} & \textcolor{mygray}{\bftab63.05\%$\pm$1.67\%} & \textcolor{mygray}{\bftab59.17\%$\pm$1.62\%} & \textcolor{mygray}{\bftab62.33\%$\pm$1.45\%}  & \textcolor{mygray}{\bftab59.89\%$\pm$1.94\%}  &\textcolor{mygray}{\bftab59.58\%$\pm$1.52\%}  &   & \textcolor{mygray}{\bftab64.34\%$\pm$1.33\%}  & \textcolor{mygray}{\bftab62.19\%$\pm$2.58\%} & \textcolor{mygray}{\bftab63.39\%$\pm$1.47\%} \\
    \hline

 \multirow{2}{*}{ \scriptsize 7} & \textcolor{black}{\bftab92.91\%$\pm$0.85\%} & \textcolor{black}{\bftab89.95\%$\pm$1.10\%}& \textcolor{black}{\bftab86.57\%$\pm$1.82\%} & \textcolor{black}{\bftab89.03\%$\pm$1.54\%}& \textcolor{black}{\bftab83.94\%$\pm$1.67\%} &   \textcolor{black}{\bftab89.69\%$\pm$0.69\%} & \textcolor{black}{\bftab91.84\%$\pm$0.68\%}  & - & \textcolor{black}{\bftab85.26\%$\pm$1.54\%} &\textcolor{black}{\bftab82.04\%$\pm$1.21\%}\\
                                       & \textcolor{mygray}{\bftab62.88\%$\pm$0.82\%} & \textcolor{mygray}{\bftab61.14\%$\pm$1.75\%} & \textcolor{mygray}{\bftab60.18\%$\pm$1.49\%} & \textcolor{mygray}{\bftab62.99\%$\pm$1.82\%}  & \textcolor{mygray}{\bftab60.16\%$\pm$1.62\%}  &\textcolor{mygray}{\bftab61.39\%$\pm$1.79\%}  & \textcolor{mygray}{\bftab64.34\%$\pm$1.33\%} &    & \textcolor{mygray}{\bftab63.14\%$\pm$1.05\%} & \textcolor{mygray}{\bftab61.79\%$\pm$1.42\%} \\
    \hline

     \multirow{2}{*}{ \scriptsize 8} & \textcolor{black}{\bftab85.45\%$\pm$0.75\%} & \textcolor{black}{\bftab82.46\%$\pm$1.60\%}& \textcolor{black}{\bftab83.78\%$\pm$1.51\%} & \textcolor{black}{\bftab82.62\%$\pm$1.43\%}& \textcolor{black}{\bftab84.87\%$\pm$0.96\%} &   \textcolor{black}{\bftab79.31\%$\pm$1.43\%} & \textcolor{black}{\bftab84.14\%$\pm$1.26\%}  & \textcolor{black}{\bftab85.26\%$\pm$1.54\%} &  &\textcolor{black}{\bftab83.54\%$\pm$0.74\%}\\
                                       & \textcolor{mygray}{\bftab60.44\%$\pm$1.42\%} & \textcolor{mygray}{\bftab60.37\%$\pm$0.61\%} & \textcolor{mygray}{\bftab59.57\%$\pm$2.13\%} & \textcolor{mygray}{\bftab61.13\%$\pm$0.83\%}  & \textcolor{mygray}{\bftab60.85\%$\pm$1.47\%}  &\textcolor{mygray}{\bftab59.90\%$\pm$1.41\%}  & \textcolor{mygray}{\bftab62.19\%$\pm$2.58\%} & \textcolor{mygray}{\bftab63.14\%$\pm$1.05\%}  &  & \textcolor{mygray}{\bftab62.47\%$\pm$1.72\%} \\
    \hline

      \multirow{2}{*}{ \scriptsize 9} & \textcolor{black}{\bftab91.87\%$\pm$0.65\%} & \textcolor{black}{\bftab87.98\%$\pm$1.77\%}& \textcolor{black}{\bftab86.82\%$\pm$1.91\%} & \textcolor{black}{\bftab87.15\%$\pm$1.23\%}& \textcolor{black}{\bftab76.98\%$\pm$1.61\%} &   \textcolor{black}{\bftab85.83\%$\pm$1.58\%} & \textcolor{black}{\bftab90.52\%$\pm$1.15\%}  & \textcolor{black}{\bftab82.04\%$\pm$1.21\%} & \textcolor{black}{\bftab83.54\%$\pm$0.74\%} &-\\
                                       & \textcolor{mygray}{\bftab64.12\%$\pm$1.35\%} & \textcolor{mygray}{\bftab61.68\%$\pm$2.08\%} & \textcolor{mygray}{\bftab59.79\%$\pm$2.05\%} & \textcolor{mygray}{\bftab62.00\%$\pm$1.37\%}  & \textcolor{mygray}{\bftab58.87\%$\pm$1.03\%}  &\textcolor{mygray}{\bftab60.32\%$\pm$1.50\%}  & \textcolor{mygray}{\bftab63.39\%$\pm$1.47\%} & \textcolor{mygray}{\bftab61.79\%$\pm$1.42\%}  & \textcolor{mygray}{\bftab62.47\%$\pm$1.72\%} &   \\
    \hline

   \bottomrule
  
   \end{tabular}
   
  \vspace{1ex}

{\raggedright ~~$^\dagger$Mean Value$\pm$Confidence Interval (0.95) \par}

\end{table*}


\setlength\tabcolsep{6pt}
\begin{table*}[tbh!]
   \caption{Classification Accuracy$^\dagger$ \\ {\scriptsize Fashion-MNIST Dataset}  \\{\scriptsize Difference between mean values: $\Delta\mu\sim \mathcal{N}(0,1)$} \\{\scriptsize \textcolor{black}{$\bullet $\, Ours\,(black)}~~~\textcolor{mygray}{ $\bullet$\,Fisher\,(gray)}}} 
   \label{tab:FMNIST}
   \tiny
   \centering
   \begin{tabular}{c|cccccccccc}
   \toprule\toprule
  
   \scriptsize \textbf{Class $\#$} &  \scriptsize 0 &  \scriptsize  1&  \scriptsize  2&  \scriptsize  3 &  \scriptsize  4 &  \scriptsize 5 & \scriptsize  6 &  \scriptsize 7 &  \scriptsize  8 &  \scriptsize  9 \\ 
     \hline

   \multirow{2}{*}{ \scriptsize 0} & - & \textcolor{black}{\bftab87.60\%$\pm$1.62\%}& \textcolor{black}{\bftab84.59\%$\pm$0.45\%} & \textcolor{black}{\bftab85.07\%$\pm$1.29\%}& \textcolor{black}{\bftab83.82\%$\pm$1.76\%} &   \textcolor{black}{\bftab90.61\%$\pm$0.62\%} & \textcolor{black}{\bftab84.01\%$\pm$0.56\%}  & \textcolor{black}{\bftab89.82\%$\pm$0.90\%} & \textcolor{black}{\bftab86.51\%$\pm$0.81\%} &\textcolor{black}{\bftab90.13\%$\pm$0.87\%}\\
                                       &  & \textcolor{mygray}{\bftab60.49\%$\pm$2.00\%} & \textcolor{mygray}{\bftab64.18\%$\pm$1.24\%} & \textcolor{mygray}{\bftab64.35\%$\pm$0.90\%}  & \textcolor{mygray}{\bftab64.79\%$\pm$1.86\%}  &\textcolor{mygray}{\bftab56.31\%$\pm$0.67\%}  & \textcolor{mygray}{\bftab58.86\%$\pm$0.98\%} & \textcolor{mygray}{\bftab64.83\%$\pm$1.40\%}  & \textcolor{mygray}{\bftab58.02\%$\pm$1.17\%} & \textcolor{mygray}{\bftab65.33\%$\pm$1.82\%} \\
    \hline   
    
     \multirow{2}{*}{ \scriptsize 1} & \textcolor{black}{\bftab87.60\%$\pm$1.62\%} & - & \textcolor{black}{\bftab90.31\%$\pm$0.92\%} & \textcolor{black}{\bftab84.60\%$\pm$0.91\%}& \textcolor{black}{\bftab92.06\%$\pm$0.54\%} &   \textcolor{black}{\bftab90.21\%$\pm$0.83\%} & \textcolor{black}{\bftab90.01\%$\pm$0.99\%}  & \textcolor{black}{\bftab91.75\%$\pm$1.25\%} & \textcolor{black}{\bftab93.11\%$\pm$0.66\%} &\textcolor{black}{\bftab93.89\%$\pm$0.77\%}\\
                                       & \textcolor{mygray}{\bftab60.49\%$\pm$2.00\%} &  & \textcolor{mygray}{\bftab59.48\%$\pm$1.77\%} & \textcolor{mygray}{\bftab59.49\%$\pm$1.13\%}  & \textcolor{mygray}{\bftab63.16\%$\pm$1.88\%}  &\textcolor{mygray}{\bftab57.27\%$\pm$0.61\%}  & \textcolor{mygray}{\bftab58.06\%$\pm$1.03\%} & \textcolor{mygray}{\bftab61.01\%$\pm$1.65\%}  & \textcolor{mygray}{\bftab57.41\%$\pm$0.73\%} & \textcolor{mygray}{\bftab62.01\%$\pm$1.59\%} \\
    \hline

 \multirow{2}{*}{ \scriptsize 2} & \textcolor{black}{\bftab84.59\%$\pm$0.45\%} & \textcolor{black}{\bftab90.31\%$\pm$0.92\%}& -& \textcolor{black}{\bftab87.43\%$\pm$1.06\%}& \textcolor{black}{\bftab76.07\%$\pm$0.70\%} &   \textcolor{black}{\bftab91.18\%$\pm$0.78\%} & \textcolor{black}{\bftab79.80\%$\pm$0.53\%}  & \textcolor{black}{\bftab91.91\%$\pm$0.65\%} & \textcolor{black}{\bftab88.39\%$\pm$0.65\%} &\textcolor{black}{\bftab91.22\%$\pm$0.88\%}\\
                                       & \textcolor{mygray}{\bftab64.18\%$\pm$1.24\%} & \textcolor{mygray}{\bftab59.48\%$\pm$1.77\%} &   & \textcolor{mygray}{\bftab62.45\%$\pm$2.39\%}  & \textcolor{mygray}{\bftab61.31\%$\pm$1.11\%}  &\textcolor{mygray}{\bftab56.34\%$\pm$0.57\%}  & \textcolor{mygray}{\bftab61.26\%$\pm$2.02\%} & \textcolor{mygray}{\bftab64.62\%$\pm$1.42\%}  & \textcolor{mygray}{\bftab58.80\%$\pm$1.27\%} & \textcolor{mygray}{\bftab64.29\%$\pm$2.06\%} \\
    \hline

 \multirow{2}{*}{ \scriptsize 3} & \textcolor{black}{\bftab85.07\%$\pm$1.29\%} & \textcolor{black}{\bftab84.60\%$\pm$0.91\%}& \textcolor{black}{\bftab87.43\%$\pm$1.06\%} & - & \textcolor{black}{\bftab87.19\%$\pm$0.64\%} &   \textcolor{black}{\bftab91.07\%$\pm$1.05\%} & \textcolor{black}{\bftab87.77\%$\pm$0.75\%}  & \textcolor{black}{\bftab90.87\%$\pm$0.80\%} & \textcolor{black}{\bftab90.11\%$\pm$0.44\%} &\textcolor{black}{\bftab91.36\%$\pm$1.19\%}\\
                                       & \textcolor{mygray}{\bftab64.35\%$\pm$0.90\%} & \textcolor{mygray}{\bftab59.49\%$\pm$1.13\%} & \textcolor{mygray}{\bftab62.45\%$\pm$2.39\%} &   & \textcolor{mygray}{\bftab64.88\%$\pm$1.43\%}  &\textcolor{mygray}{\bftab56.05\%$\pm$0.55\%}  & \textcolor{mygray}{\bftab59.23\%$\pm$1.13\%} & \textcolor{mygray}{\bftab65.03\%$\pm$1.02\%}  & \textcolor{mygray}{\bftab57.96\%$\pm$0.71\%} & \textcolor{mygray}{\bftab66.57\%$\pm$0.94\%} \\
    \hline

     \multirow{2}{*}{ \scriptsize 4} & \textcolor{black}{\bftab83.82\%$\pm$1.76\%} & \textcolor{black}{\bftab92.06\%$\pm$0.54\%}& \textcolor{black}{\bftab76.07\%$\pm$0.70\%} & \textcolor{black}{\bftab87.19\%$\pm$0.64\%}& - &   \textcolor{black}{\bftab89.86\%$\pm$0.66\%} & \textcolor{black}{\bftab78.81\%$\pm$0.79\%}  & \textcolor{black}{\bftab92.21\%$\pm$0.39\%} & \textcolor{black}{\bftab89.13\%$\pm$0.79\%} &\textcolor{black}{\bftab92.37\%$\pm$0.67\%}\\
                                       & \textcolor{mygray}{\bftab64.79\%$\pm$1.86\%} & \textcolor{mygray}{\bftab63.16\%$\pm$1.88\%} & \textcolor{mygray}{\bftab61.31\%$\pm$1.11\%} & \textcolor{mygray}{\bftab64.88\%$\pm$1.43\%}  &   &\textcolor{mygray}{\bftab55.96\%$\pm$0.65\%}  & \textcolor{mygray}{\bftab57.60\%$\pm$0.77\%} & \textcolor{mygray}{\bftab64.71\%$\pm$1.30\%}  & \textcolor{mygray}{\bftab57.05\%$\pm$1.04\%} & \textcolor{mygray}{\bftab65.87\%$\pm$1.78\%} \\
    \hline

     \multirow{2}{*}{ \scriptsize 5} & \textcolor{black}{\bftab90.61\%$\pm$0.62\%} & \textcolor{black}{\bftab90.21\%$\pm$0.83\%}& \textcolor{black}{\bftab91.18\%$\pm$0.78\%} & \textcolor{black}{\bftab91.07\%$\pm$1.05\%}& \textcolor{black}{\bftab89.86\%$\pm$0.66\%} &  - & \textcolor{black}{\bftab90.69\%$\pm$0.67\%}  & \textcolor{black}{\bftab81.82\%$\pm$0.96\%} & \textcolor{black}{\bftab88.69\%$\pm$0.91\%} &\textcolor{black}{\bftab80.41\%$\pm$1.02\%}\\
                                       & \textcolor{mygray}{\bftab56.31\%$\pm$0.67\%} & \textcolor{mygray}{\bftab57.27\%$\pm$0.61\%} & \textcolor{mygray}{\bftab56.34\%$\pm$0.57\%} & \textcolor{mygray}{\bftab56.05\%$\pm$0.55\%}  & \textcolor{mygray}{\bftab55.96\%$\pm$0.65\%}  &   & \textcolor{mygray}{\bftab56.06\%$\pm$0.88\%} & \textcolor{mygray}{\bftab55.87\%$\pm$0.61\%}  & \textcolor{mygray}{\bftab56.43\%$\pm$1.20\%} & \textcolor{mygray}{\bftab55.77\%$\pm$0.72\%} \\
    \hline

     \multirow{2}{*}{ \scriptsize 6} & \textcolor{black}{\bftab84.01\%$\pm$0.56\%} & \textcolor{black}{\bftab90.01\%$\pm$0.99\%}& \textcolor{black}{\bftab79.80\%$\pm$0.53\%} & \textcolor{black}{\bftab87.77\%$\pm$0.75\%}& \textcolor{black}{\bftab78.81\%$\pm$0.79\%} &   \textcolor{black}{\bftab90.69\%$\pm$0.67\%} & -  & \textcolor{black}{\bftab91.76\%$\pm$0.73\%} & \textcolor{black}{\bftab87.80\%$\pm$0.60\%} &\textcolor{black}{\bftab91.36\%$\pm$0.82\%}\\
                                       & \textcolor{mygray}{\bftab58.86\%$\pm$0.98\%} & \textcolor{mygray}{\bftab58.06\%$\pm$1.03\%} & \textcolor{mygray}{\bftab61.26\%$\pm$2.02\%} & \textcolor{mygray}{\bftab59.23\%$\pm$1.13\%}  & \textcolor{mygray}{\bftab57.60\%$\pm$0.77\%}  &\textcolor{mygray}{\bftab56.06\%$\pm$0.88\%}  &   & \textcolor{mygray}{\bftab60.01\%$\pm$1.46\%}  & \textcolor{mygray}{\bftab60.65\%$\pm$1.71\%} & \textcolor{mygray}{\bftab59.96\%$\pm$2.00\%} \\
    \hline

 \multirow{2}{*}{ \scriptsize 7} & \textcolor{black}{\bftab89.82\%$\pm$0.90\%} & \textcolor{black}{\bftab91.75\%$\pm$1.25\%}& \textcolor{black}{\bftab91.91\%$\pm$0.65\%} & \textcolor{black}{\bftab90.87\%$\pm$0.80\%}& \textcolor{black}{\bftab92.21\%$\pm$0.39\%} &   \textcolor{black}{\bftab81.82\%$\pm$0.96\%} & \textcolor{black}{\bftab91.76\%$\pm$0.73\%}  & - & \textcolor{black}{\bftab92.62\%$\pm$0.79\%} &\textcolor{black}{\bftab91.62\%$\pm$0.94\%}\\
                                       & \textcolor{mygray}{\bftab64.83\%$\pm$1.40\%} & \textcolor{mygray}{\bftab61.01\%$\pm$1.65\%} & \textcolor{mygray}{\bftab64.62\%$\pm$1.42\%} & \textcolor{mygray}{\bftab65.03\%$\pm$1.02\%}  & \textcolor{mygray}{\bftab64.71\%$\pm$1.30\%}  &\textcolor{mygray}{\bftab55.87\%$\pm$0.61\%}  & \textcolor{mygray}{\bftab60.01\%$\pm$1.46\%} &    & \textcolor{mygray}{\bftab58.76\%$\pm$2.50\%} & \textcolor{mygray}{\bftab66.14\%$\pm$1.43\%} \\
    \hline

     \multirow{2}{*}{ \scriptsize 8} & \textcolor{black}{\bftab86.51\%$\pm$0.81\%} & \textcolor{black}{\bftab93.11\%$\pm$0.66\%}& \textcolor{black}{\bftab88.39\%$\pm$0.65\%} & \textcolor{black}{\bftab90.11\%$\pm$0.44\%}& \textcolor{black}{\bftab89.13\%$\pm$0.79\%} &   \textcolor{black}{\bftab88.69\%$\pm$0.91\%} & \textcolor{black}{\bftab87.80\%$\pm$0.60\%}  & \textcolor{black}{\bftab92.62\%$\pm$0.79\%} &  &\textcolor{black}{\bftab89.56\%$\pm$0.42\%}\\
                                       & \textcolor{mygray}{\bftab58.02\%$\pm$1.17\%} & \textcolor{mygray}{\bftab57.41\%$\pm$0.73\%} & \textcolor{mygray}{\bftab58.80\%$\pm$1.27\%} & \textcolor{mygray}{\bftab57.96\%$\pm$0.71\%}  & \textcolor{mygray}{\bftab57.05\%$\pm$1.04\%}  &\textcolor{mygray}{\bftab56.43\%$\pm$1.20\%}  & \textcolor{mygray}{\bftab60.65\%$\pm$1.71\%} & \textcolor{mygray}{\bftab58.76\%$\pm$2.50\%}  &  & \textcolor{mygray}{\bftab58.08\%$\pm$1.30\%} \\
    \hline

      \multirow{2}{*}{ \scriptsize 9} & \textcolor{black}{\bftab90.13\%$\pm$0.87\%} & \textcolor{black}{\bftab93.89\%$\pm$0.77\%}& \textcolor{black}{\bftab91.22\%$\pm$0.88\%} & \textcolor{black}{\bftab91.36\%$\pm$1.19\%}& \textcolor{black}{\bftab92.37\%$\pm$0.67\%} &   \textcolor{black}{\bftab80.41\%$\pm$1.02\%} & \textcolor{black}{\bftab91.36\%$\pm$0.82\%}  & \textcolor{black}{\bftab91.62\%$\pm$0.94\%} & \textcolor{black}{\bftab89.56\%$\pm$0.42\%} &-\\
                                       & \textcolor{mygray}{\bftab65.33\%$\pm$1.82\%} & \textcolor{mygray}{\bftab62.01\%$\pm$1.59\%} & \textcolor{mygray}{\bftab64.29\%$\pm$2.06\%} & \textcolor{mygray}{\bftab66.57\%$\pm$0.94\%}  & \textcolor{mygray}{\bftab65.87\%$\pm$1.78\%}  &\textcolor{mygray}{\bftab55.77\%$\pm$0.72\%}  & \textcolor{mygray}{\bftab59.96\%$\pm$2.00\%} & \textcolor{mygray}{\bftab66.14\%$\pm$1.43\%}  & \textcolor{mygray}{\bftab58.08\%$\pm$1.30\%} &   \\
    \hline

   \bottomrule
  
   \end{tabular}
   
  \vspace{1ex}

 {\raggedright ~~$^\dagger$Mean Value$\pm$Confidence Interval (0.95) \par}

\end{table*}


\vspace{0.4cm}

\section{Related Work}\label{sec:rw}

Feature selection is a process of selecting $m$ features from a set $n$ features to enhance a model's performance. The source from which features are selected can be an original dataset. Alternatively, new features can be created from an original dataset via a mapping function; then, a feature selection scheme is subsequently employed to select a subset of features from this newly-created set.  The process of creating new features from original ones is referred to as feature construction. Feature construction followed by feature selection is termed feature extraction~\cite{Guyon}.

Based on the method employed to select a feature, the literature on feature selection can be organized into three categories: filters, wrappers, and embedded methods~\cite{Xue,Li}. Here we will focus on filter methods as our results  complement studies in this area.

In filter methods, features are ranked according to a given criterion. This approach is called a filter because it is used to ``\textit{filters out}’’ features that have low predictive power~\cite{Sebban}. A fairly large number of filter methods have been proposed~\cite{Guyon,Bolon}; examples include:

  \bigskip

 \textbf{Pearson's correlation coefficient.}  This approach is a linear correlation measure in which features and class labels are treated as random variables. The strength of the correlation between these two random variables is used to rank features~\cite{Arai}.
     
 \bigskip

 \textbf{Information-theoretic distances.} Similar to  Pearson's correlation coefficient approach, features and class labels are treated as random variables in information-theoretic measures~\cite{DI}. The key difference, however, between these two approaches is that information-theoretic distances can capture nonlinear dependencies between random variables.
 
 \bigskip
 
\textbf{The Kolmogorov–Smirnov test} is a hypothesis test to determine whether samples of two classes are generated by the same distribution~\cite{Pratt}. The lower the probability of the null hypothesis, the more likely a feature is beneficial for classification.
     
 \bigskip

\textbf{Relief method.} In this approach, a sample $x$ is randomly selected, without replacement, from a training set. Then, two distances, for a given feature, are measured: \textbf{1)}~$d(x,x_s)$: the distance of $x$ to its  nearest neighbor $x_s$ of the same class~~  \textbf{2)}~$d(x,x_d)$: the distance of $x$ to its  nearest neighbor $x_d$ of a different class. This process is repeated $n$ times to obtain a relevance index $J_i$ given by
\begin{gather*}
J_{i}=J_{i\text{-}1}+\frac{1}{n} \biggr(d(x,x_d)-d(x,x_s)\biggr),~i=1,2,\dots,n.
\end{gather*}

\noindent A large value of $J_i$ indicates that a feature has high relevance for classification~\cite{Kira1,kira2}.

 \medskip

     \textbf{Volume of overlap.} 
For a given feature,  this quantity measures the amount of overlap between the tails of two class-conditional distributions~\cite{Ho}. The lower the overlap, the higher a feature is ranked.

 \bigskip

   \textbf{Fisher's criterion.} This criterion is a ratio  of the between-class  $(\mu_1-\mu_0)^2$ to the within-class variance ($\sigma^2_1+\sigma^2_0$). This ratio is used in Fisher's discriminant analysis, in which high-dimensional data is projected onto a  line. The goal of this projection is to find a line on which Fisher's ratio is maximized. After which, a linear classifier is used to distinguish between classes~\cite{Fisher, Bishop}. Fisher's criterion assesses the separation between class distributions with which features are ranked. The higher the value, the better a feature is for classification.

 What sets our work apart from prior research efforts is that we develop a  filter method (Eq.~\ref{eq:divergence}) by extending  Fisher's criterion to settings where the mean values of class-conditional distributions can be severely close to each other, $\mu_1-\mu_0\approx 0$.

\clearpage

\balance

\section{Conclusion and Future Work}\label{sec:con}

A challenging problem with Fisher's criterion is that it performs poorly when mean values of class-conditional distributions are close to each other. We obtained a solution to this problem that utilizes the heteroscedasticity of data to distinguish classes from each other. This solution offers a method to extend Fisher’s criterion to domains where mean values of distributions can be problematically close to one another, $\Delta\mu\sim \mathcal{N}(0,1)$. We also described how our theoretical results can be casted into a neural network framework. Experimental results demonstrate that the techniques devised herein to solve classification problems have potential.

 The scope of this study was limited to a binary classification setting. As such, a natural progression of this work is to generalize our results to multiple classes. Additionally, the proposed extension ranks features individually according to their discriminative power. While this approach has a low computational requirement, it may perform suboptimally because there could be a synergistic interaction among features~\cite{Zhang}. A promising research direction, therefore, is to utilize said synergies to boost the performance of the proposed extension.

To conclude, the results of this paper should be of particular interest to researchers and practitioners as it addresses a practical issue: mean values of distributions may not always be far apart. One could, in practice, encounter a case when they are very close to each other.

\vspace{.7cm}

\section*{Acknowledgment}
We would like to thank Prof. Maneesh Sahani and  Prof. Shiro Ikeda  for their valuable comments and suggestions. We are grateful for the help and support provided by the Scientific Computing and Data Analysis section of the Research Support Division at OIST.

\vspace{0.8cm}

\bibliographystyle{IEEEtran}
\bibliography{MyBib.bib}

\clearpage

\appendices
\onecolumn
\section{Intuition of Assumption 2}\label{Assumption2}

\vspace{1cm}

\begin{figure}[thb!]
\begin{center}
\includegraphics[scale=0.35]{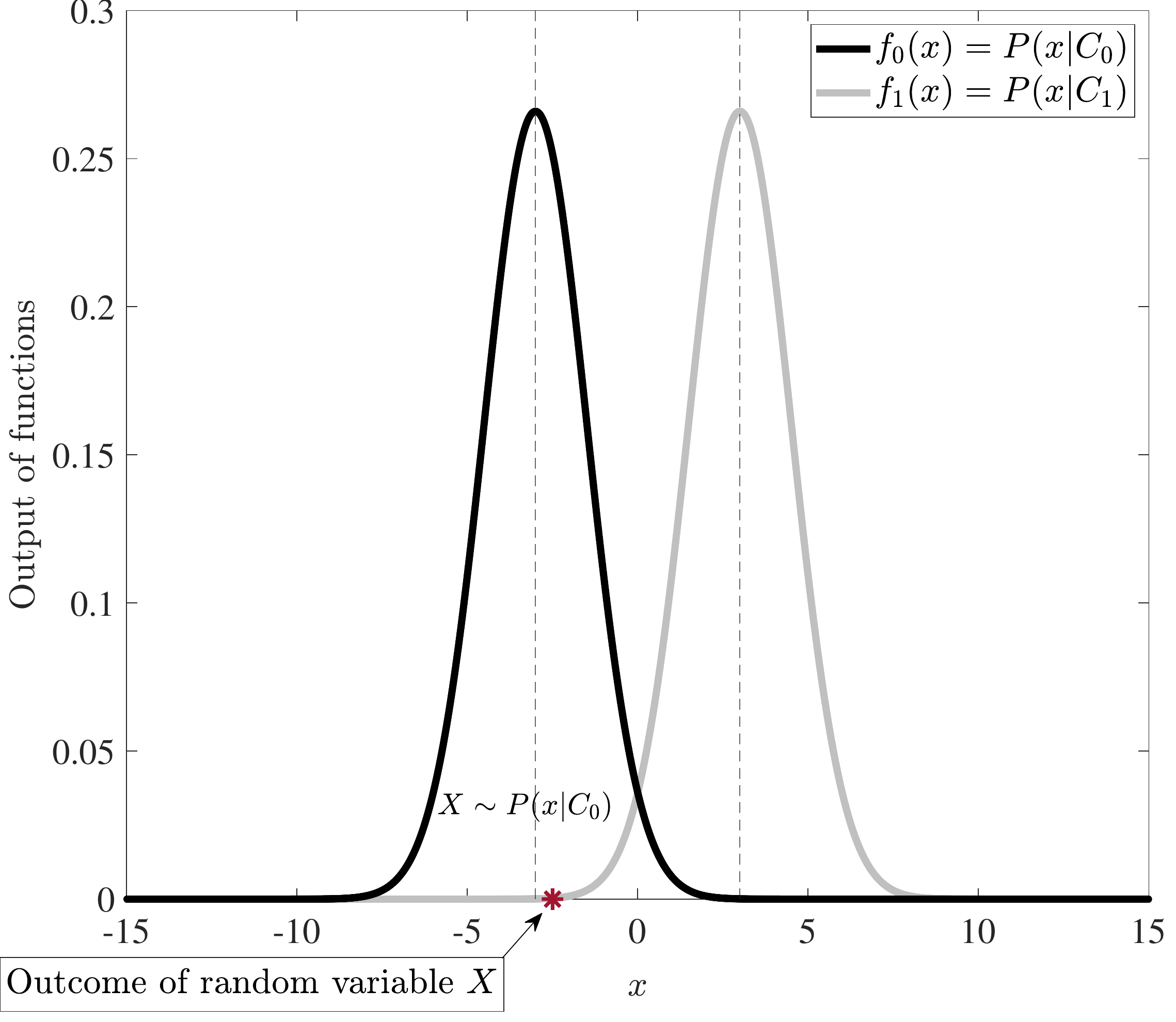}%
\end{center}
\caption{An example to illustrate Assumption 2 of Section~\ref{sec:tr}.}
\label{fig:Activation}
\end{figure}

Consider the following example shown in Fig.~\ref{fig:Activation}. Let $f_0(x)$ and $f_1(x)$ be some functions (for instance, activation functions). Moreover, for simplicity, let $P(C_0)=P(C_1) $. 

\vspace{1cm}

If $f_0(x)$ and  $f_1(x)$  are designed to have the same shape as distribution $ P(x|C_{0})$ and $ P(x|C_{1})$, respectively,  then 

\begin{align} 
\mathbb{E}_{X\sim P(x|C_{0})} [f_0(x)]> \mathbb{E}_{X\sim P(x|C_{0})} [f_{1}(x)]~~; \nonumber
\end{align} 

\bigskip

\noindent  the reason being is that an outcome, $x$, of random variable $X\sim P(x|C_{0})$ will on average fall in the region where $f_0(x)$ is near its maximum---conversely, not on average in the region where $f_1(x)$ is close to its maximum.
An equivalent example can be constructed for the pair $f_1(x)$ and $X\sim P(x|C_{1})$ as well. Lastly,  prior probabilities, $P(C_0)$ and $P(C_1)$, in inequality~\ref{eq:def} are used to factor in bias towards a class.

\clearpage

\section{Additional illustrations for proposed scheme} \label{fig:Appendix3}

\end{document}